\documentclass[11pt]{article}

\usepackage[preprint]{acl}

\usepackage{times}
\usepackage{latexsym}
\usepackage[T1]{fontenc}
\usepackage[utf8]{inputenc}
\usepackage{microtype}
\usepackage{inconsolata}
\usepackage{fvextra}
\usepackage{graphicx}
\usepackage{booktabs}
\usepackage{amsmath}
\usepackage{amssymb}
\usepackage{multirow}
\usepackage{tabularx}
\usepackage{longtable}
\usepackage{tikz}
\usepackage{pgfplots}
\usepackage{placeins}
\usetikzlibrary{arrows.meta}
\pgfplotsset{compat=1.18}

\usepackage{times}
\usepackage{latexsym}

\usepackage[T1]{fontenc}
\usepackage[utf8]{inputenc}

\usepackage{microtype}

\usepackage{inconsolata}

\usepackage{graphicx}

\title{When Retrieval Helps and Distracts: Evaluating Evidence-Generating LLMs for Biomedical Claim Verification}

\author{Pritam Deka \\
  Queen's University Belfast \\ United Kingdom \\
  \texttt{p.deka@qub.ac.uk} \\\And
  Prabhjot Singh \\
  University of Texas at Austin \\ Austin, Texas, USA \\
  \texttt{prabhjot.singh@utexas.edu} \\}

\begin{document}
\maketitle

\begin{abstract}
Biomedical fact-checking systems must do more than predict whether a claim is supported, contradicted, or unaddressed: they should also produce evidence that is faithful, complete, and useful for verification. We study this evidence-generation setting on CARE-XAI, a unified benchmark spanning five biomedical and health fact-checking sources. We compare base instruction LLMs, PubMed retrieval-augmented LLMs, fine-tuned LLMs, label-only LLMs, and biomedical encoder classifiers under a shared evaluation protocol. Biomedical classifiers remain strongest for verdict-only prediction, while fine-tuned LLMs are the strongest evidence-generating systems. PubMed retrieval is mixed: it helps PubMed-aligned sources such as PubMedQA and SciFact, but can distract models on broader public-health claims. We introduce Bio-GRACE, a gold-reference-normalized diagnostic for measuring whether retrieved evidence recovers the decision benefit of reference evidence. Bio-GRACE shows that retrieval utility is source-dependent, motivates selective retrieval, and exposes why retrieval recall and lexical evidence overlap are insufficient for biomedical fact-checking.
\end{abstract}

\section{Introduction}

Biomedical fact-checking systems must provide more than verdicts: their evidence should be faithful, complete, and useful for review. This requirement is particularly important in health applications, where fluent explanations may convert association into causation, omit uncertainty, or generalize beyond the studied population \citep{maynez2020faithfulness,ji2023hallucination}.

Retrieval-augmented generation (RAG) is often used to ground model outputs in external evidence \citep{lewis2020rag,asai2024selfrag}. However, retrieving an authoritative and topically related biomedical abstract does not guarantee that it addresses the claim being verified. This problem is especially relevant for heterogeneous benchmarks that combine scientific claims with public-health reporting and misinformation.

We investigate this issue using CARE-XAI, which contains 17,803 examples drawn from PubMedQA, SciFact, HealthVer, PUBHEALTH, and HealthFC. On its 1,752-example test set, we compare base LLMs, PubMed RAG LLMs, fine-tuned LLMs, label-only LLMs, and biomedical encoder classifiers under a shared protocol. We investigate how verdict-only and evidence-generating systems differ, when PubMed retrieval helps or distracts, and whether its decision utility can be measured relative to trusted reference evidence. %We examine three questions: how verdict-only and evidence-generating systems differ; when PubMed retrieval helps or distracts; and whether retrieval utility can be measured relative to trusted reference evidence.

%Our main contribution is Bio-GRACE, a gold-reference-normalized diagnostic that measures how much of the decision benefit provided by reference evidence is recovered by retrieved context. Our results show that biomedical classifiers remain strongest for verdict prediction, while fine-tuning is the most reliable adaptation strategy for evidence-generating LLMs. PubMed retrieval helps on PubMed-aligned sources but often distracts on broader public-health claims, motivating selective rather than always-on retrieval.

We present a systematic evaluation of evidence-generating LLMs for biomedical claim verification, comparing base, PubMed RAG, and fine-tuned systems with verdict-only LLMs and biomedical classifiers under a shared protocol. As part of this evaluation, we introduce Bio-GRACE (Biomedical Gold-Reference Assessment of Contextual Evidence), a gold-reference-normalized diagnostic that measures how much of the decision benefit provided by reference evidence is recovered by retrieved context. The results show that classifiers remain strongest for verdict prediction, while fine-tuning is the most reliable adaptation strategy for evidence-generating LLMs. PubMed retrieval helps on PubMed-aligned sources but often distracts on broader public-health claims, motivating selective rather than always-on retrieval.

\section{Related Work}

\paragraph{Biomedical and scientific fact-checking.}
FEVER, PUBHEALTH, SciFact, HealthVer, PubMedQA, and HealthFC pair claims with verdicts and supporting evidence \citep{thorne2018fever,kotonya2020pubhealth,wadden2020scifact,sarrouti2021healthver,jin2019pubmedqa,vladika2023healthfc}. MultiFC highlights cross-domain and source variation \citep{augenstein2019multifc}, while BioASQ and SciBERT provide complementary biomedical settings \citep{tsatsaronis2015bioasq,beltagy2019scibert}. CARE-XAI unifies several such resources under a common verification schema \citep{carexai2026dataset}. However, their evidence ranges from scientific abstracts to news, policy, and expert guidance, making a single PubMed retriever appropriate for some sources but mismatched to others.

\paragraph{Biomedical retrieval and evidence use.}
Health evidence extraction motivates retrieval-centred fact-checking \citep{deka2022improved,deka2022evidence,deka2023multiple}, typically combining lexical or dense retrieval with reranking \citep{robertson2009bm25,karpukhin2020dpr,nogueira2020monot5}. Recent biomedical systems also emphasize citation grounding, claim decomposition, and source retrieval \citep{ji2026medragchecker,kosprdic2026verifai,kim2025medbiorag,barone2025cer}. Rather than evaluating only plausibility or citation quality, Bio-GRACE measures how much of the decision benefit of reference evidence is recovered by retrieved PubMed context.

\paragraph{RAG evaluation and faithfulness.}
RAGAS, ARES, and RAGChecker assess relevance, faithfulness, and correctness \citep{es2023ragas,saadfalcon2023ares,ru2024ragchecker}; attribution methods separate evidential support from fluency \citep{gao2023rarr,wu2024refchecker}; and irrelevant context can degrade generation \citep{yoran2023making,zeng2025worse}. We therefore evaluate retrieval as an intervention that should move a verifier toward the correct decision.

\paragraph{Rationales and generated evidence.}
Faithful rationales should reflect the prediction process and remain grounded in evidence \citep{deyoung2020eraser,jacovi2020faithful}. Biomedical verification is stricter because fluent evidence may omit uncertainty, imply causality, or transfer findings across populations. Medical NLI is useful but vulnerable to domain artifacts \citep{romanov2018mednli,herlihy2021mednli}; we therefore evaluate generated evidence directly, using lexical overlap only as a secondary diagnostic.

\section{Task and Dataset}

We define the dataset as
\begin{equation}
\mathcal{D}=\{(x_i,y_i,e_i^\star,s_i)\}_{i=1}^{N},
\end{equation}
\begin{equation}
\mathcal{Y}=\{\textsc{Sup},\textsc{Con},\textsc{Una}\}.
\end{equation}
These labels denote supported, contradicted, and unaddressed. Here \(x_i\) is a biomedical or health claim, \(y_i\in\mathcal{Y}\) is the reference verdict, \(e_i^\star\) is reference evidence, and \(s_i\) is the source dataset. An evidence-generating verifier returns
\begin{equation}
f_\theta(x_i,c_i)\rightarrow(\hat{y}_i,\hat{e}_i,\hat{r}_i),
\end{equation}
where \(c_i\) is optional retrieved context, \(\hat{y}_i\) is the predicted verdict, \(\hat{e}_i\) generated evidence, and \(\hat{r}_i\) a short explanation.

The \textsc{Unaddressed} label is especially important. It is not simply an ``unknown'' class; it marks cases where the available evidence does not establish support or contradiction for the claim. This makes the task harder than binary biomedical entailment. A model that treats every topically related abstract as support will over-predict \textsc{Supported}; a model that treats missing direct evidence as contradiction will over-predict \textsc{Contradicted}. Good evidence generation must therefore preserve uncertainty and absence of evidence.

\begin{table}[t]
\centering
\small
\begin{tabular}{lrrrr}
\toprule
Split & Rows & Supp. & Contr. & Unaddr. \\
\midrule
Train & 14,254 & 6,840 & 4,001 & 3,413 \\
Validation & 1,797 & 841 & 565 & 391 \\
Test & 1,752 & 851 & 495 & 406 \\
\bottomrule
\end{tabular}
\caption{CARE-XAI split composition by verdict label. The test set is source-heterogeneous and label-imbalanced, motivating macro-F1 and source-stratified analysis.}
\label{tab:dataset-splits}
\end{table}

\begin{table}[t]
\centering
\small
\begin{tabular}{lrrr}
\toprule
Source & Train & Val. & Test \\
\midrule
PubMedQA & 782 & 113 & 105 \\
SciFact & 761 & 106 & 90 \\
HealthVer & 4,238 & 544 & 510 \\
PUBHEALTH & 7,866 & 964 & 974 \\
HealthFC & 607 & 70 & 73 \\
\bottomrule
\end{tabular}
\caption{CARE-XAI source composition. PubMedQA and SciFact are more directly aligned with PubMed-style source retrieval than PUBHEALTH and HealthFC.}
\label{tab:dataset-sources}
\end{table}

The source distribution in Table~\ref{tab:dataset-sources} also explains why source-stratified analysis is necessary. PUBHEALTH and HealthVer dominate the test set, while PubMedQA, SciFact, and HealthFC are smaller but methodologically distinct. A single aggregate macro-F1 can therefore hide whether a method improves scientific-abstract claims, public-health claims, or only the majority source. Throughout the paper we use aggregate results for readability and appendix tables for source-level audit.

\section{Systems}

\iffalse

\begin{figure*}[t]
\centering
\resizebox{0.98\textwidth}{!}{\input{figures/fig_methodology_pipeline}}
\caption{Evaluation workflow. CARE-XAI claims are evaluated through evidence-generating LLMs, PubMed RAG, fine-tuned LLMs, verdict-only probes, Bio-GRACE retrieval utility, and human-verification diagnostics.}
\label{fig:methodology}
\end{figure*}

\fi

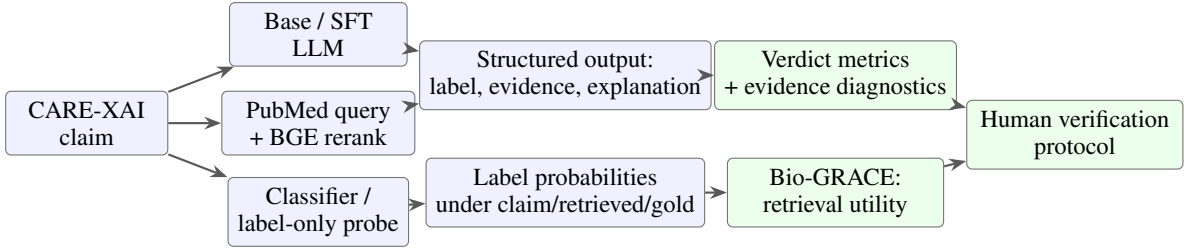
\begin{figure*}[t]
\centering
\resizebox{0.98\textwidth}{!}{\begin{tikzpicture}[
  box/.style={draw=black!55, rounded corners=2pt, align=center, minimum height=0.78cm, font=\small, fill=blue!6},
  evalbox/.style={draw=black!55, rounded corners=2pt, align=center, minimum height=0.78cm, font=\small, fill=green!8},
  arrow/.style={-{Stealth[length=2.2mm]}, thick, draw=black!65},
]
\node[box, minimum width=2.1cm] (claim) at (0,0) {CARE-XAI\\claim};
\node[box, minimum width=2.3cm] (base) at (3.0,1.15) {Base / SFT\\LLM};
\node[box, minimum width=2.5cm] (rag) at (3.0,0) {PubMed query\\+ BGE rerank};
\node[box, minimum width=2.3cm] (clf) at (3.0,-1.15) {Classifier /\\label-only probe};
\node[box, minimum width=2.6cm] (gen) at (6.2,0.62) {Structured output:\\label, evidence, explanation};
\node[box, minimum width=2.6cm] (prob) at (6.2,-0.9) {Label probabilities\\under claim/retrieved/gold};
\node[evalbox, minimum width=2.8cm] (metrics) at (9.7,0.62) {Verdict metrics\\+ evidence diagnostics};
\node[evalbox, minimum width=2.8cm] (bio) at (9.7,-0.9) {Bio-GRACE:\\retrieval utility};
\node[evalbox, minimum width=2.8cm] (human) at (12.8,-0.15) {Human verification\\protocol};

\draw[arrow] (claim) -- (base);
\draw[arrow] (claim) -- (rag);
\draw[arrow] (claim) -- (clf);
\draw[arrow] (base) -- (gen);
\draw[arrow] (rag) -- (gen);
\draw[arrow] (clf) -- (prob);
\draw[arrow] (gen) -- (metrics);
\draw[arrow] (prob) -- (bio);
\draw[arrow] (metrics) -- (human);
\draw[arrow] (bio) -- (human);
\end{tikzpicture}}
\caption{Overview of the evaluation framework. CARE-XAI claims are processed by base, PubMed RAG, and fine-tuned LLMs for verdict and evidence generation. Verdict-only probes, Bio-GRACE, and human evaluation assess evidence use and retrieval utility.}
\label{fig:methodology}
\end{figure*}

%\paragraph{Prompting strategies.}
%We evaluate four strategies: zero-shot, chain-of-thought few-shot, PICO zero-shot, and PICO few-shot. All strategies use the same three-way verdict definitions and structured JSON output containing a label, a synthesised evidence passage, and an explanation. RAG prompts additionally instruct models to prioritise retrieved evidence when it is sufficient and to reason conservatively when it is weak or conflicting. Complete prompt templates and few-shot examples are provided in Appendix~\ref{app:prompts}.

\paragraph{Base LLMs and prompting.}
We evaluate instruction-tuned Gemma, Qwen, and Gemini-family systems \citep{gemma2024open,gemma2025gemma3,yang2024qwen25,team2024gemini15} using zero-shot, chain-of-thought few-shot, PICO zero-shot, and PICO few-shot prompting. All strategies use the same three-way verdict definitions and structured JSON output containing a label, a synthesised evidence passage, and an explanation. RAG variants additionally instruct models to prioritise retrieved evidence when sufficient and reason conservatively when it is weak or conflicting. Outputs are parsed into the three corresponding fields. Complete templates and demonstrations are provided in Appendix~\ref{app:prompts}.

%\paragraph{Base LLMs.}
%We evaluate instruction-tuned Gemma, Qwen, and Gemini-family systems under four prompting strategies \citep{Abd2026Gemma4T,yang2024qwen25,team2023gemini,comanici2025gemini}. Outputs are parsed into verdict, evidence, and explanation fields.

\paragraph{PubMed RAG LLMs.}
Following prior biomedical evidence-retrieval work \citep{deka2022improved}, each claim is rewritten into a PubMed-oriented query. PubMed candidates are retrieved through Entrez, titles and abstracts are fetched, and BGE-M3 reranks claim-document pairs \citep{chen2024bgem3}. The top contexts are inserted into the prompt. The system is evaluated both as a generator and as a retrieval pipeline.

To make model comparisons controlled, we separate retrieval from answer generation. We first build a frozen retrieval cache for the full test set, then use the same cached contexts for every answer model, prompt, and ablation. Each cache row stores the generated query, retrieved PMIDs, titles, abstracts, publication types, original PubMed ranks, BGE reranker scores, and selected top-\(k\) context. This avoids confounding answer-model comparisons with different PubMed calls, timestamps, or reranker behavior. It also lets us attribute failures more precisely: if the cache lacks useful evidence, the error is primarily retrieval-side; if useful evidence is present but the answer is wrong or unsupported, the error is generation-side.

\paragraph{Fine-tuned LLMs.}
Fine-tuned LLMs use supervised instruction tuning on the CARE-XAI training split. Each training instance is formatted as a chat example with a system instruction, a user message containing the claim and CARE-XAI reference evidence, and an assistant response containing structured JSON with \texttt{label}, \texttt{evidence\_text}, and \texttt{explanation}. During loss computation, only assistant tokens are trained. At test time, the original non-RAG inference scripts provide only the claim and ask the adapted model to generate all three fields. Thus, these runs test whether evidence-conditioned supervision transfers to claim-only evidence generation; they are not a matched claim-only training protocol.

We fine-tune Gemma and Qwen variants with LoRA \citep{hu2022lora} through Unsloth. This parameter-efficient adaptation follows instruction-tuning and low-memory fine-tuning practice \citep{ouyang2022instructgpt,dettmers2023qlora}. The runs use one training epoch, batch size 1 with gradient accumulation 4, learning rate \(2\times10^{-4}\), linear scheduling, warmup of 100 steps, and checkpoint/evaluation every 200 steps. The maximum sequence length is 15k tokens for Gemma and 16k for Qwen; LoRA rank is scaled by model size. We ran one fine-tuning instance per model configuration with the trainer's fixed seed 3407 rather than a repeated-seed grid. These systems are evaluated without PubMed retrieval, so their gains reflect supervised adaptation to CARE-XAI evidence formatting and label semantics rather than test-time retrieval.

\paragraph{Classifiers and label-only LLMs.}
Biomedical encoders provide verdict-only probes. They cannot generate evidence, so they are not complete fact-checking systems. We additionally run evidence-conditioned classifier and label-only LLM diagnostics using claim-only, retrieved-evidence, and gold-evidence inputs. Gold evidence is an oracle diagnostic, not a deployable baseline.

\begin{table}[t]
\centering
\small
\resizebox{\columnwidth}{!}{%
\begin{tabular}{lll}
\toprule
Family & Input & Output \\
\midrule
Base LLM & claim & label, evidence, explanation \\
RAG LLM & claim + PubMed context & label, evidence, explanation \\
Fine-tuned LLM & claim & label, evidence, explanation \\
Label-only LLM & claim/context/evidence & label only \\
Encoder classifier & claim/context/evidence & label probabilities \\
\bottomrule
\end{tabular}
}
\caption{System families and output contracts. Only the first three families generate evidence; classifiers and label-only LLMs are diagnostic verdict probes.}
\label{tab:system-contracts}
\end{table}

Table~\ref{tab:system-contracts} summarizes the system contracts. This distinction keeps comparisons fair. Evidence-generating LLMs must produce a structured answer that can be inspected, while classifiers are optimized only for labels. We therefore use classifiers to probe verdict predictability and evidence usefulness, but not as replacements for evidence-generating systems.

\paragraph{Quality gates and manifest policy.}
The evaluation treats model failures as part of the empirical picture. Runs with missing files, incomplete row counts, invalid labels, empty evidence, or template-copy behavior remain in the experiment manifest. They are not silently discarded; instead, paper-facing quality metrics are set to null when outputs cannot support a valid comparison. %Mistral and Ministral outputs are excluded from paper-facing summaries because they are outside the final model scope, while raw directories are preserved. 
Duplicate non-RAG baseline directories are collapsed when corresponding prediction files are byte-identical.

\paragraph{Decoding and alignment.}
All complete files are aligned to the 1,752-row CARE-XAI test split by row index and claim occurrence. This matters because some sources contain duplicate claim identifiers. LLM outputs are parsed into normalized labels and evidence fields; invalid labels count against valid-label coverage. Classifier rows are marked as verdict-only so evidence coverage and evidence-overlap metrics are not interpreted as missing generated evidence.

\section{Evaluation}

\paragraph{Verdict and output quality.}
The primary verdict metric is macro-F1, with accuracy, balanced accuracy, MCC, valid-label coverage, parse coverage, and evidence coverage as diagnostics. Duplicate claim identifiers are aligned by row index/occurrence. Incomplete runs remain in the manifest with null quality metrics rather than being silently removed.

Macro-F1 is the main endpoint because the test set is label-imbalanced and because minority classes, especially \textsc{Contradicted} and \textsc{Unaddressed}, are central for fact-checking. Accuracy is still reported because it is intuitive, but it can hide systems that over-predict the majority supported class. Evidence coverage is reported separately from verdict metrics because a system can produce a valid label with empty or template-like evidence. ROUGE and BERTScore are retained only as secondary lexical/semantic similarity diagnostics \citep{lin2004rouge,zhang2020bertscore}.

\paragraph{Generated explanations.}
Generated explanations are retained for transparency but not treated as headline evidence-faithfulness scores. Appendix~\ref{app:explanation-diag} reports only an internal explanation--verdict consistency check; generated evidence remains the primary verifiable artifact. An anonymized artifact provides the evaluation scripts and paper-facing result files.\footnote{\url{https://anonymous.4open.science/r/care-xai-173E/}}

\paragraph{Retrieval diagnostics.}
The frozen PubMed cache is evaluated with query success, MRR, Recall@1/5/10, and nDCG@10 for examples with known source PMIDs. These metrics measure whether source evidence is retrieved, but not whether the retrieved text improves verification.

This separation is important for biomedical RAG. A retrieved abstract may contain correct biomedical facts and still be irrelevant to the exact verification decision. Conversely, a document may not match the original source PMID but may still contain evidence that helps the verifier. We therefore report retrieval metrics as diagnostics and use Bio-GRACE to measure downstream decision utility.

\begin{table}[t]
\centering
\small
\begin{tabular}{lrrrr}
\toprule
Metric & Value & \multicolumn{3}{c}{Scope} \\
\midrule
Rows & 1,752 & \multicolumn{3}{c}{all test rows} \\
Query success & 0.871 & \multicolumn{3}{c}{all test rows} \\
MRR & 0.212 & \multicolumn{3}{c}{known PMIDs} \\
Recall@1 & 0.205 & \multicolumn{3}{c}{known PMIDs} \\
Recall@10 & 0.224 & \multicolumn{3}{c}{known PMIDs} \\
nDCG@10 & 0.214 & \multicolumn{3}{c}{known PMIDs} \\
\bottomrule
\end{tabular}
\caption{Frozen PubMed retrieval-cache diagnostics. Retrieval succeeds for most claims, but source-document recall is low over rows with known source PMIDs.}
\label{tab:retrieval-main}
\end{table}

\begin{figure}[t]
\centering
\includegraphics[width=\columnwidth]{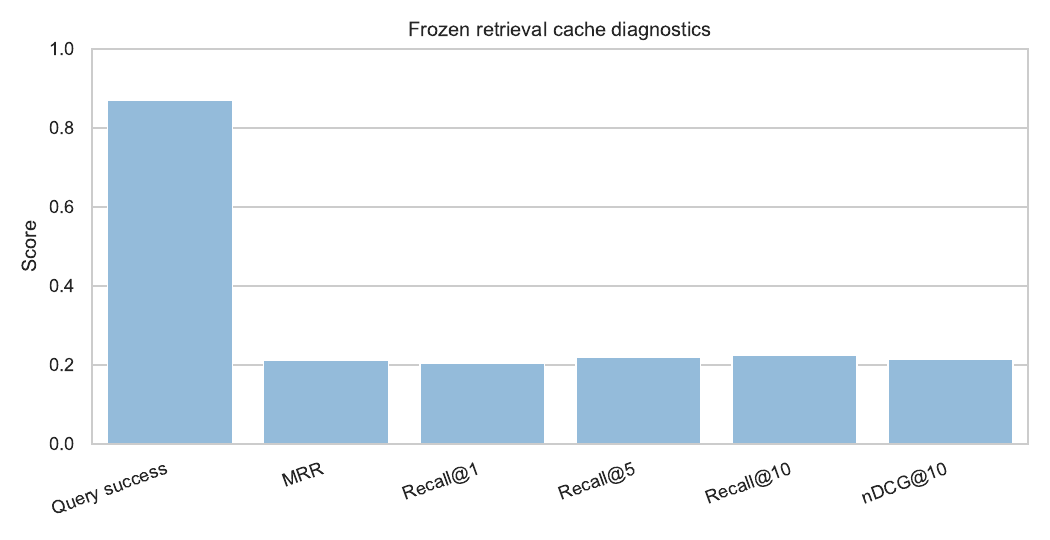}
\caption{Frozen PubMed retrieval diagnostics by source. Query success is high, but source-PMID recovery is uneven and often weak for public-health sources.}
\label{fig:retrieval-diagnostics-main}
\end{figure}

Figure~\ref{fig:retrieval-diagnostics-main} visualizes why retrieval metrics are useful but insufficient. PubMed can return many abstracts for a claim without retrieving the source evidence that determines the CARE-XAI verdict. This motivates separating retrieval success, downstream verdict accuracy, and Bio-GRACE utility instead of treating retrieved context as automatically beneficial.

\paragraph{Bio-GRACE.}
For a verifier that outputs a probability \(p_i(y)\), define the claim-only, retrieved, and gold-evidence true-label probabilities as \(p_i^C(y_i)\), \(p_i^R(y_i)\), and \(p_i^G(y_i)\). The oracle utility is
\begin{equation}
U_i^\star = p_i^G(y_i)-p_i^C(y_i),
\end{equation}
and retrieved utility is
\begin{equation}
U_i^R = p_i^R(y_i)-p_i^C(y_i).
\end{equation}
For evidence-sensitive examples \(\mathcal{H}=\{i:U_i^\star>0\}\), Bio-GRACE utility recovery is
\begin{equation}
\mathrm{UR}=\frac{1}{|\mathcal{H}|}\sum_{i\in\mathcal{H}}\mathrm{clip}\left(\frac{U_i^R}{U_i^\star+\epsilon},-1,1\right).
\end{equation}
UR is positive when retrieved context recovers reference-evidence benefit, near zero when retrieval adds little, and negative when retrieval moves probability mass away from the correct label.

We also report class-conditional and source-conditional variants in the appendix. The source-conditioned view is essential because a negative aggregate UR can result from a mixture of helpful retrieval on PubMed-aligned sources and harmful retrieval on public-health sources. We use Bio-GRACE as an evaluation diagnostic, not as a training objective.

\paragraph{Uncertainty diagnostics.}
For classifiers, we compute predictive entropy and margin:
\begin{equation}
H_i=-\sum_{y\in\mathcal{Y}}p_i(y)\log p_i(y),
\end{equation}
\begin{equation}
m_i=p_i^{(1)}-p_i^{(2)},
\end{equation}
where \(p_i^{(1)}\) and \(p_i^{(2)}\) are the largest and second-largest class probabilities. For LLMs without stored token probabilities, we use vote disagreement across available model outputs as a proxy:
\begin{equation}
\hat{p}_{i,c}^{vote}=\frac{1}{|\mathcal{M}_i|}\sum_{m\in\mathcal{M}_i}\mathbf{1}[\hat{y}_i^{(m)}=c].
\end{equation}
These uncertainty measures are not used to select final systems in the main results. They are included to show whether confidence and disagreement can help identify cases where evidence should be reviewed or retrieval should be gated. This follows calibration, ensemble, verbalized-uncertainty, and selective-prediction work \citep{guo2017calibration,lakshminarayanan2017simple,kadavath2022know,lin2022uncertainty,geifman2017selective,angelopoulos2021uncertainty}.

%\paragraph{Human verification.}
%Human annotation is ongoing. We prepared a simplified blinded packet with 100 final items and five calibration items for two biotech annotators, covering verdict correctness, evidence support, usefulness, and safety concern. The final analysis will report agreement with Cohen's kappa and Krippendorff's alpha \citep{cohen1960,krippendorff2018}.

%\paragraph{Leakage and sensitivity.}
%Because CARE-XAI merges evidence-centered sources, we audit overlap across the official split and report evidence-safe sensitivity results. This subset is not a replacement benchmark; it tests whether conclusions depend on repeated evidence. Matched systems use paired resampling and corrected comparisons \citep{efron1979bootstrap,mcnemar1947,holm1979}.

\section{Results}

\begin{table*}[t]
\centering
\small
\begin{tabular}{lllrrrrrr}
\toprule
Family & Model & Prompt & Macro-F1 & Bal. Acc. & MCC & Acc. & Valid & Evid. \\
\midrule
Non-RAG LLM & Qwen3.5-27B & zero-shot & 0.450 & 0.442 & 0.182 & 0.483 & 0.998 & 0.998 \\
PubMed RAG LLM & Qwen3.5-9B & zero-shot & 0.431 & 0.421 & 0.155 & 0.461 & 0.999 & 1.000 \\
Fine-tuned LLM & Gemma-4-31B & PICO few-shot & 0.483 & 0.476 & 0.247 & 0.538 & 1.000 & 1.000 \\
Classifier & BioBERT-base & claim-only & 0.560 & 0.555 & 0.359 & 0.604 & 1.000 & -- \\
\bottomrule
\end{tabular}
\caption{Best complete run in each evaluated family. Classifiers are verdict-only probes and do not generate evidence.}
\label{tab:main-results}
\end{table*}

\begin{table}[t]
\centering
\small
\resizebox{\columnwidth}{!}{%
\begin{tabular}{lrrrr}
\toprule
Regime & Runs & Mean F1 & Best F1 & Mean Acc. \\
\midrule
Biomedical classifiers & 9 & 0.547 & 0.560 & 0.587 \\
Non-RAG LLM & 44 & 0.394 & 0.450 & 0.408 \\
PubMed RAG LLM & 43 & 0.385 & 0.431 & 0.423 \\
Fine-tuned LLM & 13 & 0.447 & 0.483 & 0.503 \\
\bottomrule
\end{tabular}
}
\caption{Aggregate summary over complete runs. Fine-tuning is more reliable than always-on PubMed RAG for evidence-generating LLMs.}
\label{tab:regime-summary}
\end{table}

Table~\ref{tab:main-results} shows the core tension. Biomedical classifiers are strongest for verdict-only prediction, but they do not generate evidence. Among evidence-generating systems, the best fine-tuned LLM reaches 0.483 macro-F1, outperforming the best base and RAG LLMs. Table~\ref{tab:regime-summary} shows the aggregate pattern: fine-tuning improves LLMs more consistently than retrieval.

The completed Qwen3.5-27B grid reinforces the mixed retrieval result: RAG lowers macro-F1 in three of four matched prompts, while PICO few-shot improves only from 0.417 to 0.421. It does not change the best-system rankings or the central conclusions.

This should not be read as evidence that classifiers solve the task. Instead, it separates two capabilities that are often conflated: selecting the correct verdict and producing inspectable evidence. The gap between classifiers and evidence-generating LLMs suggests that evidence generation imposes an additional burden beyond label prediction. For biomedical applications, this burden is desirable to measure because a correct label with no evidence is difficult to audit, while plausible evidence with a wrong or unsupported label can be harmful.

The LLM results also show that model scale alone is not the central story. Some larger systems fail because of output schema drift, invalid labels, or evidence templates that do not correspond to the claim. Smaller or fine-tuned systems can be more reliable when the output contract is stable. This is why the manifest retains incomplete and low-coverage runs: excluding them would overstate the maturity of evidence-generating biomedical verification.

\begin{figure}[t]
\centering
\resizebox{\columnwidth}{!}{\begin{tikzpicture}[x=11.0cm,y=0.54cm]
\scriptsize
\draw[gray!35] (0,-0.75) -- (0,6.2);
\foreach \x in {0.2,0.4,0.6} {
  \draw[gray!18] (\x,-0.75) -- (\x,6.2);
  \node[font=\tiny,anchor=north] at (\x,-0.75) {\x};
}
\node[anchor=east,font=\tiny] at (-0.012,0) {Gemma-4-E4B};
\filldraw[fill=teal!55,draw=teal!80!black] (0,-0.24) rectangle (0.3272,-0.05);
\filldraw[fill=orange!70,draw=orange!80!black] (0,-0.04) rectangle (0.3881,0.15);
\filldraw[fill=purple!55,draw=purple!80!black] (0,0.16) rectangle (0.4761,0.35);
\node[anchor=east,font=\tiny] at (-0.012,1) {Gemma-4-31B};
\filldraw[fill=teal!55,draw=teal!80!black] (0,0.76) rectangle (0.4413,0.95);
\filldraw[fill=orange!70,draw=orange!80!black] (0,0.96) rectangle (0.4090,1.15);
\filldraw[fill=purple!55,draw=purple!80!black] (0,1.16) rectangle (0.4834,1.35);
\node[anchor=east,font=\tiny] at (-0.012,2) {Qwen3.5-4B};
\filldraw[fill=teal!55,draw=teal!80!black] (0,1.76) rectangle (0.4176,1.95);
\filldraw[fill=orange!70,draw=orange!80!black] (0,1.96) rectangle (0.3822,2.15);
\filldraw[fill=purple!55,draw=purple!80!black] (0,2.16) rectangle (0.4353,2.35);
\node[anchor=east,font=\tiny] at (-0.012,3) {Qwen3.5-35B-A3B};
\filldraw[fill=teal!55,draw=teal!80!black] (0,2.76) rectangle (0.4436,2.95);
\filldraw[fill=orange!70,draw=orange!80!black] (0,2.96) rectangle (0.4000,3.15);
\filldraw[fill=purple!55,draw=purple!80!black] (0,3.16) rectangle (0.4814,3.35);
\node[anchor=east,font=\tiny] at (-0.012,4) {Qwen3.6-27B};
\filldraw[fill=teal!55,draw=teal!80!black] (0,3.76) rectangle (0.4338,3.95);
\filldraw[fill=orange!70,draw=orange!80!black] (0,3.96) rectangle (0.4107,4.15);
\filldraw[fill=purple!55,draw=purple!80!black] (0,4.16) rectangle (0.4779,4.35);
\node[anchor=east,font=\tiny] at (-0.012,5) {Qwen3.6-35B-A3B};
\filldraw[fill=teal!55,draw=teal!80!black] (0,4.76) rectangle (0.4366,4.95);
\filldraw[fill=orange!70,draw=orange!80!black] (0,4.96) rectangle (0.4221,5.15);
\filldraw[fill=purple!55,draw=purple!80!black] (0,5.16) rectangle (0.4742,5.35);

\filldraw[fill=teal!55,draw=teal!80!black] (0.00,5.85) rectangle (0.025,6.05);
\node[anchor=west,font=\tiny] at (0.028,5.95) {Base};
\filldraw[fill=orange!70,draw=orange!80!black] (0.13,5.85) rectangle (0.155,6.05);
\node[anchor=west,font=\tiny] at (0.158,5.95) {RAG};
\filldraw[fill=purple!55,draw=purple!80!black] (0.25,5.85) rectangle (0.275,6.05);
\node[anchor=west,font=\tiny] at (0.278,5.95) {Fine-tuned};
\node[font=\scriptsize] at (0.31,-1.35) {Best macro-F1};
\end{tikzpicture}}
\caption{Best macro-F1 for LLMs with complete base, RAG, and fine-tuned triples. Fine-tuning usually improves evidence-generating LLMs; RAG is mixed.}
\label{fig:all-models}
\end{figure}
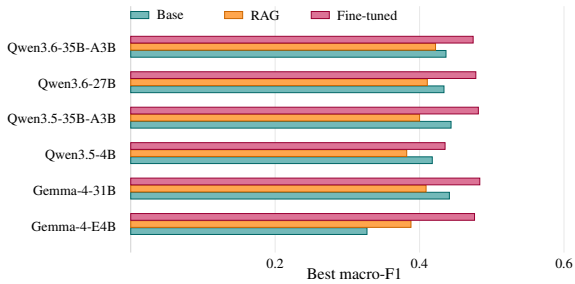

\subsection{Evidence Conditioning}

\begin{table}[t]
\centering
\small
\resizebox{\columnwidth}{!}{%
\begin{tabular}{llrr}
\toprule
System & Input & Runs & Macro-F1 \\
\midrule
Classifier & Claim only & 9 & 0.547 \\
Classifier & Retrieved PubMed & 9 & 0.386 \\
Classifier & Gold evidence & 9 & 0.692 \\
Label-only LLM & Claim only & 3 & 0.331 \\
Label-only LLM & Retrieved PubMed & 3 & 0.343 \\
Label-only LLM & Gold evidence & 3 & 0.648 \\
\bottomrule
\end{tabular}
}
\caption{Evidence-conditioning diagnostics. Gold evidence substantially improves verdict prediction, while retrieved PubMed context is much weaker and can hurt encoder probes.}
\label{tab:evidence-conditioning}
\end{table}

Table~\ref{tab:evidence-conditioning} confirms that reference evidence is decision-useful: gold evidence sharply improves both supervised encoders and label-only LLMs. Retrieved PubMed evidence is not equivalent to gold evidence. It slightly improves label-only LLMs but degrades classifier probes, consistent with retrieval noise.

\begin{figure}[t]
\centering
\includegraphics[width=\columnwidth]{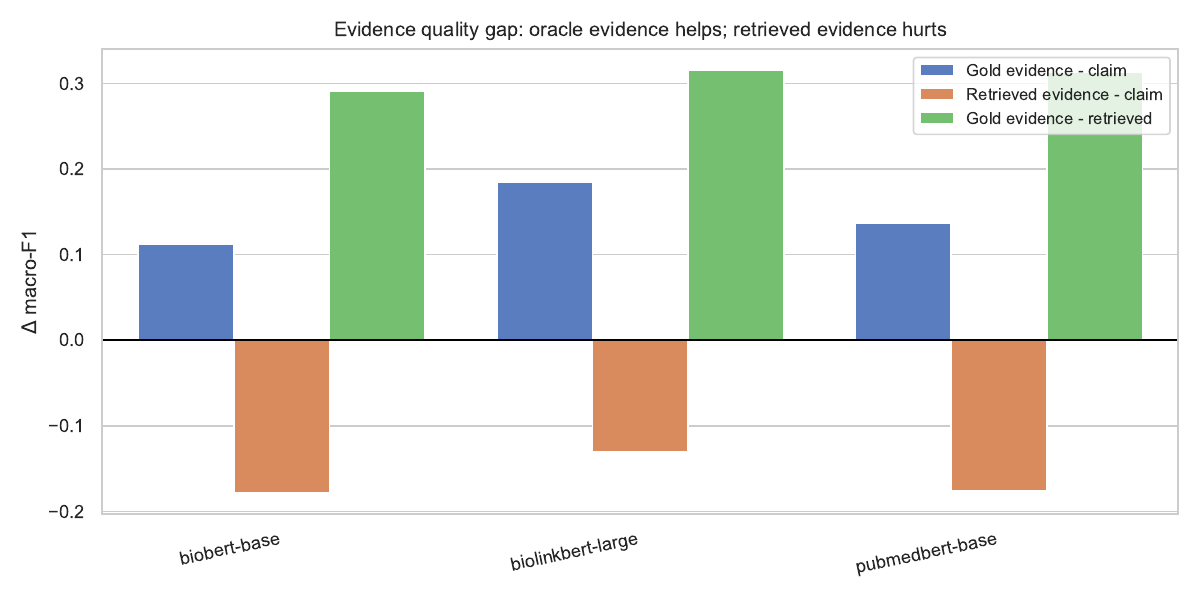}
\caption{Evidence-conditioning gaps. Gold evidence acts as an oracle diagnostic; retrieved PubMed context does not close the gap to reference evidence.}
\label{fig:evidence-gaps-main}
\end{figure}

The oracle gold-evidence setting is not deployable, but it is methodologically useful. It tells us that the task is evidence-sensitive: when high-quality evidence is supplied, both neural encoders and LLMs become substantially better verdict predictors. The failure of retrieved evidence to close this gap indicates that the bottleneck is not simply whether models can use evidence, but whether retrieval supplies the right kind of evidence.

This result also clarifies the role of evidence generation. If gold evidence improves label-only systems, then evidence contains information that models can use. If retrieved evidence does not produce comparable gains, then retrieval is the weak link. If evidence-generating LLMs still trail label-only or classifier probes, then producing evidence and a verdict together remains harder than classifying from supplied evidence. These three observations motivate evaluating retrieval, generation, and verdict prediction as separate components rather than reporting one end-to-end score.

\subsection{Retrieval Utility}

\begin{table}[t]
\centering
\small
\resizebox{\columnwidth}{!}{%
\begin{tabular}{lrrrrr}
\toprule
Model & UR & CW-UR & NRI & Distract & Rescue \\
\midrule
BioBERT & -0.110 & 0.135 & -0.147 & 0.282 & 0.135 \\
BioLinkBERT & -0.151 & 0.103 & -0.125 & 0.275 & 0.150 \\
PubMedBERT & -0.204 & 0.082 & -0.123 & 0.278 & 0.155 \\
\bottomrule
\end{tabular}
}
\caption{Bio-GRACE diagnostics. Negative UR and NRI show that always-on retrieval often distracts; positive CW-UR shows that retrieved evidence can help evidence-sensitive cases.}
\label{tab:biograce-main}
\end{table}

\begin{table}[t]
\centering
\small
\resizebox{\columnwidth}{!}{%
\begin{tabular}{lrrrr}
\toprule
Source & UR & Hit@10 & Base F1 & Ret. F1 \\
\midrule
PubMedQA & 0.676 & 0.895 & 0.118 & 0.502 \\
SciFact & 0.435 & 0.000 & 0.291 & 0.486 \\
HealthVer & -0.030 & 0.064 & 0.453 & 0.408 \\
HealthFC & -0.222 & 0.000 & 0.335 & 0.471 \\
PUBHEALTH & -0.378 & 0.048 & 0.257 & 0.224 \\
\bottomrule
\end{tabular}
}
\caption{Source-level Bio-GRACE and label-only LLM retrieval effects. PubMedQA and SciFact show positive retrieval utility; PUBHEALTH is the dominant negative-utility source.}
\label{tab:source-utility}
\end{table}

\begin{figure}[t]
\centering
\resizebox{\columnwidth}{!}{\begin{tikzpicture}
\begin{axis}[
  ybar,
  width=0.92\textwidth,
  height=5.4cm,
  ylabel={Utility score},
  symbolic x coords={PubMedQA,SciFact,HealthVer,HealthFC,PUBHEALTH},
  xtick=data,
  x tick label style={rotate=25,anchor=east,font=\small},
  ymin=-0.45,
  ymax=0.85,
  bar width=10pt,
  legend style={at={(0.5,1.03)},anchor=south,legend columns=2,font=\small},
  grid=major,
  grid style={draw=gray!18},
]
\addplot+[fill=teal!55,draw=teal!80!black] coordinates {(PubMedQA,0.661) (SciFact,0.424) (HealthVer,-0.026) (HealthFC,-0.211) (PUBHEALTH,-0.355)};
\addplot+[fill=orange!55,draw=orange!80!black] coordinates {(PubMedQA,0.759) (SciFact,0.558) (HealthVer,0.090) (HealthFC,0.028) (PUBHEALTH,-0.013)};
\legend{UR,CW-UR}
\end{axis}
\end{tikzpicture}}
\caption{Bio-GRACE utility by source. PubMedQA and SciFact benefit from PubMed retrieval, while PUBHEALTH and HealthFC show negative retrieval utility.}
\label{fig:biograce-source-main}
\end{figure}
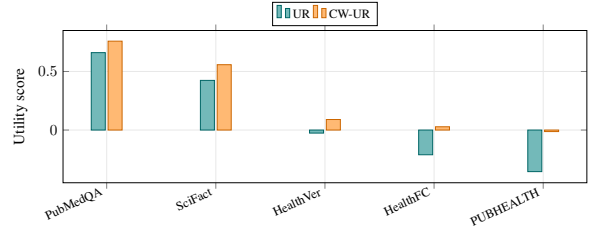

Bio-GRACE explains why RAG is mixed. Table~\ref{tab:biograce-main} shows negative average utility recovery for all three encoder verifiers. Retrieval sometimes rescues wrong predictions, but it more often distracts correct claim-only predictions. Table~\ref{tab:source-utility} shows that this is not uniform: PubMed-aligned sources benefit, while broader public-health claims often do not.

The source-level pattern is intuitive but important. PubMedQA and SciFact often ask about scientific abstracts or biomedical study claims, making PubMed retrieval a closer match to the evidence need. PUBHEALTH and HealthFC include public-health and misinformation claims whose verification may require news context, guidelines, or careful interpretation rather than direct abstract matching. A biomedical retrieval source can therefore be authoritative but source-mismatched.

The retrieval metrics in Table~\ref{tab:retrieval-main} help explain the ceiling. Query success is high, but known source-document recall remains low. This means many claims receive some PubMed context, but not necessarily the decisive source context. In a generation setting, such context can be dangerous because it gives the model vocabulary and apparent evidence for a nearby biomedical topic. The resulting answer may be more fluent and more cited, yet less faithful to the claim.

\subsection{Selective Retrieval Diagnostics}

\begin{table}[t]
\centering
\small
\resizebox{\columnwidth}{!}{%
\begin{tabular}{llrr}
\toprule
Family & Condition & Runs & Macro-F1 \\
\midrule
Classifier & claim-only & 9 & 0.547 \\
Classifier & always-retrieved & 9 & 0.386 \\
Classifier & source-router & 9 & 0.567 \\
LLM label-only & claim-only & 3 & 0.331 \\
LLM label-only & always-retrieved & 3 & 0.343 \\
LLM label-only & source-router & 3 & 0.381 \\
\bottomrule
\end{tabular}
}
\caption{No-new-inference routing simulation. The source router uses retrieval for sources with positive Bio-GRACE UR and claim-only prediction otherwise.}
\label{tab:routing}
\end{table}

The routing simulation in Table~\ref{tab:routing} is not a deployed model and does not use a validation-trained threshold. It tests whether the Bio-GRACE source signal identifies where retrieval should be used. The result motivates selective biomedical RAG: retrieval should be gated rather than applied as a blanket intervention.

The modest source-router gains should be interpreted conservatively. Because the router is a no-new-inference diagnostic, it cannot prove that a deployable system would learn the same boundary. It does, however, demonstrate that retrieval utility is structured rather than random. A future system could combine retrieval confidence, source metadata, claim type, and classifier uncertainty to decide whether to retrieve, abstain, or request human review.

\section{Leakage, NLI, and Human Verification}

Because CARE-XAI unifies multiple evidence-centered sources, we audit overlap. The official split contains exact claim overlap in 9 test rows, exact evidence overlap in 588 test rows, 591 rows removed by a stricter evidence-group-safe filter, and 360 embedding near-duplicate pairs at cosine threshold 0.95. Evidence-safe results preserve the qualitative conclusions: classifiers remain strongest for verdict-only prediction, fine-tuning remains more reliable than always-on RAG, and retrieval remains source-dependent.

\begin{table}[t]
\centering
\small
\resizebox{\columnwidth}{!}{%
\begin{tabular}{lrrr}
\toprule
Regime & Full F1 & Safe F1 & $\Delta$ \\
\midrule
Classifier & 0.547 & 0.529 & +0.018 \\
Base LLM & 0.394 & 0.356 & +0.038 \\
RAG LLM & 0.385 & 0.361 & +0.025 \\
Fine-tuned LLM & 0.448 & 0.441 & +0.006 \\
\bottomrule
\end{tabular}
}
\caption{Leakage sensitivity by regime. Safe F1 removes rows in the evidence-group-safe filter. The central retrieval and fine-tuning conclusions remain stable.}
\label{tab:leakage-main}
\end{table}

Directional NLI uses reference evidence as premise and generated evidence as hypothesis. Because results vary sharply by scorer, it is a sensitivity diagnostic rather than a replacement for human faithfulness assessment.

\begin{table}[t]
\centering
\small
\resizebox{\columnwidth}{!}{%
\begin{tabular}{lrrr}
\toprule
NLI scorer & Ent. & Contr. & DFS \\
\midrule
BioLinkBERT-MedNLI & 0.672 & 0.234 & 0.395 \\
PubMedBERT-MNLI-MedNLI & 0.517 & 0.417 & 0.304 \\
DeBERTa-v3-ZS-NLI & 0.245 & -- & 0.149 \\
RoBERTa-large-MNLI & 0.076 & 0.211 & 0.045 \\
\bottomrule
\end{tabular}
}
\caption{Directional NLI evidence diagnostics. Entailment rates vary substantially by scorer, so NLI is used as sensitivity analysis rather than a definitive faithfulness metric.}
\label{tab:nli-main}
\end{table}

\paragraph{Human verification.}
We evaluated 100 blinded evidence-generating outputs across base, RAG, and fine-tuned regimes using two biotech annotators (master's students) and two independent LLM judges (GPT-5.6-Sol and Claude Opus-4.8). Human--human agreement was modest: Cohen's $\kappa$ was 0.265 for verdict correctness, 0.391 for evidence support, 0.155 for usefulness, and 0.250 for safety. On the subsets with exact human agreement, 18.8\% (13/69) had a correct verdict, 19.6\% (11/56) contained supported or partially supported evidence, and 20.8\% (10/48) were useful or partly useful; 21.0\% (13/62) raised a safety concern, with no agreed major concern. The two LLM judges showed markedly different agreement with human consensus: $\kappa$ ranged from 0.107--0.653 for verdict correctness and evidence support, but reached 0.645 and 0.657 for usefulness. Four-rater Krippendorff's $\alpha$ remained low (0.105--0.276). We therefore use human consensus as the primary assessment and treat LLM judgments as sensitivity diagnostics, not as adjudication or ground truth.

\section{Discussion}

Biomedical evidence generation is not reducible to verdict classification. Verdict-only classifiers establish how predictable the CARE-XAI labels are, but they do not produce the reviewable evidence required in a fact-checking workflow. Evidence-generating LLMs face a harder joint problem: they must select a verdict, identify the relevant support relation, communicate it clearly, and avoid introducing unsupported biomedical details. The results suggest that fine-tuning helps mainly by teaching the dataset's label semantics and structured output contract. It improves evidence-generating systems more consistently than retrieval, although it does not itself guarantee faithful evidence.

The retrieval results reveal a source-matching problem rather than a simple failure of PubMed. PubMed is appropriate for claims whose evidence is expressed in biomedical abstracts, but CARE-XAI also contains public-health reporting and misinformation-oriented claims whose decisive context may not be recoverable through an abstract-focused query. In such cases, retrieved text can look authoritative and improve citation appearance while moving the verifier away from the correct verdict. Bio-GRACE captures this distinction by measuring how much of the decision benefit provided by trusted reference evidence is recovered by retrieved context. The source-level variation and routing simulation therefore support selective retrieval: systems should consider retrieval confidence, source compatibility, model disagreement, and missing evidence before accepting retrieved context or escalating a case for human review.

\section{Conclusion}

This study asks whether LLMs can generate useful biomedical fact-checking evidence, rather than only predict a verdict. Across CARE-XAI, classifiers remain stronger at verdict prediction, while fine-tuning most reliably improves evidence-generating LLMs. PubMed RAG helps when abstracts match a claim's evidence need but distracts otherwise. Bio-GRACE quantifies this source-dependent utility by comparing retrieved with trusted evidence, motivating selective rather than always-on retrieval. Biomedical verification should therefore treat evidence as a first-class, decision-useful, auditable, and safety-critical output.

\clearpage

%\section*{Limitations}

%Bio-GRACE measures how retrieved context changes biomedical encoders, not intrinsic truth. Gold evidence is evaluation-only and routing is a no-new-inference diagnostic, not a validation-trained system. Human annotation is ongoing; CARE-XAI is useful but not leakage-free, and PubMed is not sufficient for every heterogeneous claim.

%\section*{Ethical Considerations}

%This study is not medical advice; generated evidence can be incomplete or harmful without expert review.

\section*{Limitations}

Bio-GRACE measures how retrieved context changes the predictions of supervised biomedical verifiers; it does not measure the intrinsic truth or clinical validity of retrieved evidence. Its results may therefore depend on verifier calibration and dataset-specific decision boundaries. Reference evidence is used only for evaluation and may itself be incomplete or heterogeneous. The proposed source router is a no-new-inference diagnostic rather than a validation-trained deployable system. 

Our retrieval experiments are limited to PubMed and one retrieval pipeline, so the findings should not be interpreted as a general failure of RAG. Some few-shot conditions used non-identical demonstrations across RAG and non-RAG settings, so matched zero-shot comparisons provide the cleanest estimate of retrieval effects. Broader sources, including clinical guidelines, public-health agencies, and reputable news or policy documents, may better support some CARE-XAI claims. CARE-XAI is also source-imbalanced and contains evidence overlap across splits, although leakage-filtered analyses preserve the main conclusions. Finally, the human evaluation covers 100 outputs and two biotech annotators, with modest inter-annotator agreement; it provides an initial assessment rather than clinical validation. NLI and LLM-judge results are retained only as sensitivity diagnostics because they vary across evaluators.

\section*{Ethical Considerations}

This work evaluates research systems and does not provide medical advice or support autonomous clinical decisions. Generated evidence may omit uncertainty, introduce unsupported details, or transform associations into causal claims. Retrieved documents may also be authoritative but irrelevant to the specific claim, creating a risk of persuasive yet misleading outputs. Such systems should therefore expose their sources, preserve uncertainty, and require qualified human review before use in health-related settings.

The study uses existing benchmark data and does not involve clinical deployment. We reviewed the source datasets' documented licenses and use restrictions before redistribution; because CARE-XAI combines sources with different terms, downstream users must retain source-level attribution and comply with the most restrictive applicable terms. The two annotators were master's-level biotech students recruited voluntarily through an academic collaborator in India. Their participation was unpaid, they provided informed consent, and they evaluated model outputs rather than personal or patient data. No institutional ethics review was obtained. Human judgments are used only to evaluate system outputs, and LLM-based judges are treated as sensitivity diagnostics rather than substitutes for biomedical expertise. The proposed retrieval-routing analysis should likewise not be interpreted as a safety mechanism without prospective validation.

\bibliography{custom}

\clearpage
\appendix

\section{Appendix Overview}

The main paper is self-contained. This appendix provides detailed result tables, diagnostics, and protocol information that are useful for reproducibility and reviewer audit but not required for the main narrative.

The appendix is organized around four goals. First, it makes the experiment manifest auditable by listing complete, incomplete, and missing runs rather than only successful outputs. Second, it expands the retrieval and Bio-GRACE analysis to show why retrieval utility differs by source. Third, it reports auxiliary diagnostics such as leakage sensitivity, NLI behavior, explanation consistency, lexical overlap, calibration, uncertainty, and routing headroom, that support but do not replace the main conclusions. Fourth, it documents the human-verification protocol so that the eventual manual evaluation can be connected to the automatic evidence diagnostics reported here.

\section{Prompt Templates}
\label{app:prompts}

We evaluate four prompting strategies in both non-RAG and RAG
settings: zero-shot, chain-of-thought few-shot, PICO zero-shot, and
PICO few-shot. Dynamic fields are represented by angle-bracketed
placeholders.

\subsection{Shared Output Schema}

\begin{Verbatim}[
  frame=single,
  breaklines=true,
  breakanywhere=true,
  fontsize=\small
]
{
  "label": "<SUPPORTED|CONTRADICTED|UNADDRESSED>",
  "evidence_text": "<synthesised evidence passage, 2-5 sentences>",
  "explanation": "<reasoning explanation, 3-6 sentences>"
}
\end{Verbatim}

\subsection{Non-RAG Prompts}

\subsubsection{Zero-Shot}

\begin{Verbatim}[
  frame=single,
  breaklines=true,
  breakanywhere=true,
  fontsize=\small
]
You are a medical evidence analyst.

TASK:
Evaluate the following health claim against your knowledge of the
peer-reviewed medical literature. Determine whether the claim is:

- SUPPORTED — the evidence clearly supports the claim
- CONTRADICTED — the evidence contradicts or refutes the claim
- UNADDRESSED — the evidence is insufficient or does not address
  this claim

Then generate a concise evidence passage summarising the relevant
medical evidence, and provide an explanation of your reasoning.

HEALTH CLAIM:
<CLAIM>

Respond ONLY with a JSON object using this exact schema:
<OUTPUT_SCHEMA>
\end{Verbatim}

\subsubsection{Chain-of-Thought Few-Shot}

\begin{Verbatim}[
  frame=single,
  breaklines=true,
  breakanywhere=true,
  fontsize=\small
]
You are a medical evidence analyst trained in evidence-based medicine.

TASK:
Evaluate a health claim by first reasoning through the evidence
step-by-step (chain-of-thought), then output your structured verdict.

Verdicts:
- SUPPORTED — evidence clearly supports the claim
- CONTRADICTED — evidence refutes the claim
- UNADDRESSED — evidence is insufficient or silent

EXAMPLES:
<NON_RAG_FEW_SHOT_EXAMPLES>

--- Now evaluate the following ---
Claim: <CLAIM>

First, reason through the evidence step-by-step. Then respond ONLY
with the JSON object using this exact schema:
<OUTPUT_SCHEMA>
\end{Verbatim}

\subsubsection{PICO Zero-Shot}

\begin{Verbatim}[
  frame=single,
  breaklines=true,
  breakanywhere=true,
  fontsize=\small
]
You are a medical evidence analyst trained in Evidence-Based Medicine
(EBM).

TASK:
Evaluate the health claim below using the structured PICO framework
(Population, Intervention, Comparison, Outcome).

STEP 1 — PICO DECOMPOSITION:
Identify:
P: Population
I: Intervention
C: Comparison
O: Outcome

STEP 2 — EVIDENCE APPRAISAL:
Assess study design, strength, and limitations.

STEP 3 — VERDICT:
SUPPORTED / CONTRADICTED / UNADDRESSED

HEALTH CLAIM:
<CLAIM>

Respond ONLY with JSON:
<OUTPUT_SCHEMA>
\end{Verbatim}

\subsubsection{PICO Few-Shot}

\begin{Verbatim}[
  frame=single,
  breaklines=true,
  breakanywhere=true,
  fontsize=\small
]
You are a medical evidence analyst trained in Evidence-Based Medicine
(EBM).

TASK:
Evaluate the health claim using the PICO framework and structured
reasoning.

STEP 1 — PICO:
Identify Population, Intervention, Comparison, Outcome.

STEP 2 — EVIDENCE APPRAISAL:
Assess study type, strength, and limitations.

STEP 3 — VERDICT:
SUPPORTED / CONTRADICTED / UNADDRESSED

EXAMPLES:
<NON_RAG_FEW_SHOT_EXAMPLES>

--- Now evaluate ---
Claim: <CLAIM>

Respond ONLY with JSON:
<OUTPUT_SCHEMA>
\end{Verbatim}

\subsection{RAG Prompts}

Retrieved PubMed records are supplied as PMID, title, and abstract
blocks. If retrieval returns no documents, the prompt states:
\texttt{No retrieved biomedical evidence available.}

\subsubsection{Zero-Shot RAG}

\begin{Verbatim}[
  frame=single,
  breaklines=true,
  breakanywhere=true,
  fontsize=\small
]
You are a medical evidence analyst.

TASK:
Evaluate the following health claim against the retrieved peer-reviewed
biomedical literature below. Determine whether the claim is:

- SUPPORTED — the evidence clearly supports the claim
- CONTRADICTED — the evidence contradicts or refutes the claim
- UNADDRESSED — the evidence is insufficient or does not address
  this claim

When retrieved evidence is sufficient, prioritise it over prior
knowledge.

Then generate a concise evidence passage summarising the relevant
medical evidence, and provide an explanation of your reasoning.

HEALTH CLAIM:
<CLAIM>

RETRIEVED BIOMEDICAL LITERATURE:
<RETRIEVED_DOCUMENTS>

Respond ONLY with a JSON object using this exact schema:
<OUTPUT_SCHEMA>
\end{Verbatim}

\subsubsection{Chain-of-Thought Few-Shot RAG}

\begin{Verbatim}[
  frame=single,
  breaklines=true,
  breakanywhere=true,
  fontsize=\small
]
You are a medical evidence analyst trained in evidence-based medicine.

TASK:
Evaluate a health claim against the retrieved biomedical literature
by reasoning through the evidence step-by-step.

Verdicts:
- SUPPORTED
- CONTRADICTED
- UNADDRESSED

When retrieved evidence is sufficient, prioritise it over prior
knowledge.

EXAMPLES:
<RAG_FEW_SHOT_EXAMPLES>

--- Retrieved Evidence ---
<RETRIEVED_DOCUMENTS>

--- Now evaluate ---
Claim: <CLAIM>

Respond ONLY with JSON:
<OUTPUT_SCHEMA>
\end{Verbatim}

\subsubsection{PICO Zero-Shot RAG}

\begin{Verbatim}[
  frame=single,
  breaklines=true,
  breakanywhere=true,
  fontsize=\small
]
You are a medical evidence analyst trained in Evidence-Based Medicine
(EBM).

TASK:
Evaluate the health claim using the retrieved biomedical literature
and the structured PICO framework.

When retrieved evidence is sufficient, prioritise it over prior
knowledge.

STEP 1 — PICO DECOMPOSITION:
Identify Population, Intervention, Comparison, Outcome.

STEP 2 — EVIDENCE APPRAISAL:
Assess study design, strength, and limitations.

STEP 3 — VERDICT:
SUPPORTED / CONTRADICTED / UNADDRESSED

HEALTH CLAIM:
<CLAIM>

RETRIEVED BIOMEDICAL LITERATURE:
<RETRIEVED_DOCUMENTS>

Respond ONLY with JSON:
<OUTPUT_SCHEMA>
\end{Verbatim}

\subsubsection{PICO Few-Shot RAG}

\begin{Verbatim}[
  frame=single,
  breaklines=true,
  breakanywhere=true,
  fontsize=\small
]
You are a medical evidence analyst trained in Evidence-Based Medicine
(EBM).

TASK:
Evaluate the health claim using the retrieved biomedical literature,
PICO framework, and structured reasoning.

When retrieved evidence is sufficient, prioritise it over prior
knowledge.

STEP 1 — PICO
STEP 2 — EVIDENCE APPRAISAL
STEP 3 — VERDICT

EXAMPLES:
<RAG_FEW_SHOT_EXAMPLES>

RETRIEVED BIOMEDICAL LITERATURE:
<RETRIEVED_DOCUMENTS>

Claim: <CLAIM>

Respond ONLY with JSON:
<OUTPUT_SCHEMA>
\end{Verbatim}

\subsection{Non-RAG Few-Shot Demonstrations}
\label{app:non-rag-demonstrations}

The following demonstrations are inserted into both non-RAG few-shot
prompts. The heading is \texttt{Chain-of-Thought Reasoning} in the
chain-of-thought prompt and \texttt{PICO + Reasoning} in the PICO
prompt.

\subsubsection{Non-RAG Demonstration 1}

\begin{Verbatim}[
  frame=single,
  breaklines=true,
  breakanywhere=true,
  fontsize=\small
]
--- Example 1 ---

Claim:
Risk factors for major depression during midlife among women with and
without prior major depression are the same.

Reasoning:
Step 1 — Identify PICO elements: Population = community sample of women
aged 42–52 enrolled in the Study of Women's Health Across the Nation;
Intervention/Exposure = lifetime psychiatric history, menopausal
status, vasomotor symptoms, psychosocial factors; Comparison = women
with vs. without prior MDD history at baseline; Outcome = first-onset
vs. recurrent major depressive disorder (MDD) over 13 annual
follow-ups.

Step 2 — Appraise evidence: A longitudinal cohort study (n=443) used
Cox multivariable analyses separately for women with and without prior
MDD. Women without lifetime MDD had a substantially lower risk (28%)
than those with prior MDD (59%), and their risk profiles were
qualitatively different — health conditions and vasomotor symptoms
drove first-onset, while menopausal transition and prior anxiety
disorder drove recurrence.

Step 3 — Conclude: The claim that risk factors are the same is directly
contradicted by the study findings, which demonstrate distinct risk
profiles for first-onset vs. recurrent MDD.

Output:
{
  "label": "CONTRADICTED",
  "evidence_text": "A 13-year longitudinal cohort study of 443 midlife
  women (aged 42–52) found that risk factors for first-lifetime-onset
  MDD differed from those for recurrent MDD. Health conditions and
  vasomotor symptoms were primary drivers of first-onset MDD, whereas
  peri/post-menopausal status, psychological symptoms, and prior
  anxiety disorder were predominant risk factors for recurrence.
  Women with no prior MDD history had a substantially lower midlife
  risk (28%) compared to those with prior MDD (59%).",
  "explanation": "The claim is CONTRADICTED. A large prospective cohort
  study directly compared risk profiles for first-onset vs. recurrent
  MDD in midlife women and found them to be meaningfully different.
  First-onset MDD was associated with physical health conditions and
  vasomotor symptoms, while recurrence was driven by menopausal
  transition and prior anxiety history. These distinct profiles
  indicate that screening and prevention strategies should differ
  depending on a woman's MDD history."
}
\end{Verbatim}

\subsubsection{Non-RAG Demonstration 2}

\begin{Verbatim}[
  frame=single,
  breaklines=true,
  breakanywhere=true,
  fontsize=\small
]
--- Example 2 ---

Claim:
The leukocyte count at presentation correlates with the severity of
injury in blunt trauma patients.

Reasoning:
Step 1 — PICO: Population = blunt trauma patients admitted to an
emergency department (n=713); Intervention/Exposure = white blood cell
(WBC) count at presentation; Comparison = established injury severity
scores (ISS, GCS, RTS); Outcome = correlation between WBC and injury
severity.

Step 2 — Appraise evidence: A retrospective study of 713 patients
found WBC positively correlated with ISS and negatively correlated
with GCS and RTS — both statistically significant. WBC is objective,
quick, and easy to obtain compared to ISS (complex) or GCS/RTS
(subjective).

Step 3 — Conclude: The evidence directly supports the claim. WBC count
is a useful adjunct measure of injury severity.

Output:
{
  "label": "SUPPORTED",
  "evidence_text": "A retrospective study of 713 blunt trauma patients
  evaluated the relationship between white blood cell (WBC) count at
  presentation and established injury severity measures (ISS, GCS,
  RTS). Statistical analysis revealed that WBC count correlated
  positively with ISS and negatively with GCS and RTS, supporting its
  use as an objective, rapid adjunct indicator of injury severity in
  emergency settings.",
  "explanation": "The claim is SUPPORTED. A retrospective cohort study
  of 713 blunt trauma patients demonstrated statistically significant
  correlations between WBC count at presentation and all three
  standard injury severity scores. Because WBC is faster and more
  objective than ISS and less subjective than GCS or RTS, this
  evidence supports its clinical utility as an adjunct severity
  indicator, though the retrospective single-centre design limits
  generalisability."
}
\end{Verbatim}

\subsubsection{Non-RAG Demonstration 3}

\begin{Verbatim}[
  frame=single,
  breaklines=true,
  breakanywhere=true,
  fontsize=\small
]
--- Example 3 ---

Claim:
French lay people and health professionals uniformly agree on the
acceptability of breaching confidentiality to protect a patient's
wife from a sexually transmitted disease.

Reasoning:
Step 1 — PICO: Population = 144 lay people, 10 psychologists, and
7 physicians in France; Intervention = physician breaching patient
confidentiality to protect spouse from STD; Comparison = across
48 scenario combinations varying disease severity, time, patient
intent, protective behaviour intent, and expert consultation;
Outcome = acceptability ratings (scale 0–22).

Step 2 — Appraise evidence: The study found marked disagreement between
groups — lay people rated acceptability highest (mean 11.76),
psychologists intermediate (9.28), and physicians lowest (2.90).
Within lay participants, cluster analysis revealed four distinct
subgroups ranging from "always acceptable" to "never acceptable".
Acceptability was most influenced by the patient's stated intentions.

Step 3 — Conclude: The evidence shows substantial disagreement, not
uniform agreement.

Output:
{
  "label": "UNADDRESSED",
  "evidence_text": "A French study presented 48 hypothetical
  breach-of-confidentiality scenarios to 144 lay people,
  10 psychologists, and 7 physicians. Acceptability ratings differed
  substantially across groups.",
  "explanation": "The claim is UNADDRESSED. The study concerns ethical
  attitudes rather than clinical evidence outcomes."
}
\end{Verbatim}

\subsection{RAG Few-Shot Demonstrations}
\label{app:rag-demonstrations}

The RAG prompt file uses shortened versions of the demonstrations. As
above, the reasoning heading depends on whether the chain-of-thought or
PICO few-shot prompt is used.

\subsubsection{RAG Demonstration 1}

\begin{Verbatim}[
  frame=single,
  breaklines=true,
  breakanywhere=true,
  fontsize=\small
]
--- Example 1 ---

Claim:
Risk factors for major depression during midlife among women with and
without prior major depression are the same.

Reasoning:
Step 1 - Identify PICO elements: Population = community sample of women
aged 42–52 enrolled in the Study of Women's Health Across the Nation;
Intervention/Exposure = lifetime psychiatric history, menopausal
status, vasomotor symptoms, psychosocial factors; Comparison = women
with vs. without prior MDD history at baseline; Outcome = first-onset
vs. recurrent major depressive disorder (MDD) over 13 annual
follow-ups.

Step 2 - Appraise evidence: A longitudinal cohort study (n=443) used
Cox multivariable analyses separately for women with and without prior
MDD. Women without lifetime MDD had a substantially lower risk (28%)
than those with prior MDD (59%), and their risk profiles were
qualitatively different.

Step 3 - Conclude: The claim is contradicted.

Output:
{
  "label": "CONTRADICTED",
  "evidence_text": "A 13-year longitudinal cohort study found distinct
  risk profiles for first-onset versus recurrent MDD.",
  "explanation": "The claim is CONTRADICTED because the evidence
  directly shows different risk factors."
}
\end{Verbatim}

\subsubsection{RAG Demonstration 2}

\begin{Verbatim}[
  frame=single,
  breaklines=true,
  breakanywhere=true,
  fontsize=\small
]
--- Example 2 ---

Claim:
The leukocyte count at presentation correlates with the severity of
injury in blunt trauma patients.

Reasoning:
Step 1 - PICO.
Step 2 - Appraise evidence.
Step 3 - Conclude.

Output:
{
  "label": "SUPPORTED",
  "evidence_text": "A retrospective study of 713 blunt trauma patients
  found significant correlations.",
  "explanation": "The claim is SUPPORTED by direct evidence."
}
\end{Verbatim}

\subsubsection{RAG Demonstration 3}

\begin{Verbatim}[
  frame=single,
  breaklines=true,
  breakanywhere=true,
  fontsize=\small
]
--- Example 3 ---

Claim:
French lay people and health professionals uniformly agree on the
acceptability of breaching confidentiality.

Reasoning:
Step 1 — PICO.
Step 2 — Appraise evidence.
Step 3 — Conclude.

Output:
{
  "label": "UNADDRESSED",
  "evidence_text": "The retrieved study concerns ethical attitudes
  rather than clinical outcomes.",
  "explanation": "The claim is UNADDRESSED."
}
\end{Verbatim}

\section{Full Experiment Tables}

This section reports the complete paper-facing runs and the quality-gated outputs retained in the manifest. Table~\ref{tab:app-complete-results} lists every complete run, Table~\ref{tab:app-failures} records excluded outputs and their failure reasons, and Table~\ref{tab:app-manifest} summarizes manifest status. %These tables are intentionally long: EACL permits unlimited appendix material, and the goal is to make result inclusion and exclusion auditable.

\begin{table*}[!t]
\centering
\small
\setlength{\tabcolsep}{1.5pt}
\begin{tabular*}{\textwidth}{@{\extracolsep{\fill}}lllrrrrrrrr@{}}
\toprule
Reg. & Model & Prompt & Seed & F1 & Acc. & Bal. & MCC & Valid & Evid. & R-L \\
\midrule
baseline & BioBERT-base & claim-only & 13 & 0.555 & 0.600 & 0.552 & 0.354 & 1.000 & 0.000 & -- \\
baseline & BioBERT-base & claim-only & 42 & 0.560 & 0.604 & 0.555 & 0.359 & 1.000 & 0.000 & -- \\
baseline & BioBERT-base & claim-only & 2026 & 0.550 & 0.573 & 0.556 & 0.343 & 1.000 & 0.000 & -- \\
baseline & BioLinkBERT-large & claim-only & 13 & 0.541 & 0.588 & 0.538 & 0.333 & 1.000 & 0.000 & -- \\
baseline & BioLinkBERT-large & claim-only & 42 & 0.538 & 0.587 & 0.535 & 0.329 & 1.000 & 0.000 & -- \\
baseline & BioLinkBERT-large & claim-only & 2026 & 0.542 & 0.583 & 0.541 & 0.333 & 1.000 & 0.000 & -- \\
baseline & PubMedBERT-base & claim-only & 13 & 0.536 & 0.575 & 0.536 & 0.325 & 1.000 & 0.000 & -- \\
baseline & PubMedBERT-base & claim-only & 42 & 0.543 & 0.579 & 0.543 & 0.334 & 1.000 & 0.000 & -- \\
baseline & PubMedBERT-base & claim-only & 2026 & 0.558 & 0.590 & 0.560 & 0.354 & 1.000 & 0.000 & -- \\
base & Gemma-4-12B & cot-fewshot & 0 & 0.375 & 0.385 & 0.381 & 0.091 & 0.999 & 0.999 & 0.076 \\
base & Gemma-4-12B & pico-fewshot & 0 & 0.347 & 0.355 & 0.369 & 0.074 & 1.000 & 1.000 & 0.076 \\
base & Gemma-4-12B & pico-zero-shot & 0 & 0.369 & 0.388 & 0.384 & 0.098 & 0.998 & 0.999 & 0.075 \\
base & Gemma-4-12B & zero-shot & 0 & 0.411 & 0.431 & 0.406 & 0.132 & 1.000 & 1.000 & 0.078 \\
base & Gemma-4-26B-A4B & cot-fewshot & 0 & 0.425 & 0.434 & 0.421 & 0.154 & 0.998 & 0.998 & 0.078 \\
base & Gemma-4-26B-A4B & pico-fewshot & 0 & 0.395 & 0.394 & 0.399 & 0.113 & 0.999 & 0.999 & 0.078 \\
base & Gemma-4-26B-A4B & pico-zero-shot & 0 & 0.390 & 0.396 & 0.393 & 0.107 & 0.997 & 1.000 & 0.078 \\
base & Gemma-4-26B-A4B & zero-shot & 0 & 0.402 & 0.416 & 0.400 & 0.119 & 1.000 & 1.000 & 0.080 \\
base & Gemma-4-31B & cot-fewshot & 0 & 0.441 & 0.468 & 0.431 & 0.169 & 0.997 & 0.997 & 0.081 \\
base & Gemma-4-31B & pico-fewshot & 0 & 0.415 & 0.414 & 0.407 & 0.124 & 1.000 & 1.000 & 0.081 \\
base & Gemma-4-31B & pico-zero-shot & 0 & 0.414 & 0.422 & 0.405 & 0.124 & 1.000 & 1.000 & 0.075 \\
base & Gemma-4-31B & zero-shot & 0 & 0.417 & 0.429 & 0.405 & 0.124 & 1.000 & 1.000 & 0.079 \\
base & Gemma-4-E2B & cot-fewshot & 0 & 0.365 & 0.368 & 0.391 & 0.099 & 0.998 & 0.998 & 0.074 \\
base & Gemma-4-E2B & pico-fewshot & 0 & 0.363 & 0.371 & 0.365 & 0.069 & 0.999 & 0.999 & 0.073 \\
base & Gemma-4-E2B & pico-zero-shot & 0 & 0.337 & 0.336 & 0.356 & 0.047 & 0.999 & 0.999 & 0.077 \\
base & Gemma-4-E2B & zero-shot & 0 & 0.334 & 0.338 & 0.359 & 0.058 & 0.999 & 0.999 & 0.082 \\
base & Gemma-4-E4B & pico-fewshot & 0 & 0.327 & 0.336 & 0.363 & 0.069 & 1.000 & 1.000 & 0.074 \\
base & Qwen3.5-27B & cot-fewshot & 0 & 0.450 & 0.466 & 0.441 & 0.174 & 1.000 & 1.000 & 0.092 \\
base & Qwen3.5-27B & pico-fewshot & 0 & 0.417 & 0.418 & 0.408 & 0.126 & 1.000 & 1.000 & 0.089 \\
base & Qwen3.5-27B & pico-zero-shot & 0 & 0.426 & 0.434 & 0.414 & 0.135 & 1.000 & 1.000 & 0.091 \\
base & Qwen3.5-27B & zero-shot & 0 & 0.450 & 0.483 & 0.442 & 0.182 & 0.998 & 0.998 & 0.093 \\
base & Qwen3.5-2B & cot-fewshot & 0 & 0.335 & 0.382 & 0.358 & 0.056 & 0.991 & 0.991 & 0.092 \\
base & Qwen3.5-2B & pico-fewshot & 0 & 0.333 & 0.355 & 0.338 & 0.023 & 0.991 & 0.991 & 0.091 \\
base & Qwen3.5-2B & pico-zero-shot & 0 & 0.345 & 0.340 & 0.391 & 0.093 & 0.997 & 0.997 & 0.099 \\
base & Qwen3.5-2B & zero-shot & 0 & 0.318 & 0.348 & 0.357 & 0.054 & 0.999 & 0.999 & 0.097 \\
base & Qwen3.5-35B-A3B & cot-fewshot & 0 & 0.416 & 0.442 & 0.403 & 0.121 & 0.997 & 0.991 & 0.095 \\
base & Qwen3.5-35B-A3B & pico-fewshot & 0 & 0.375 & 0.392 & 0.368 & 0.066 & 0.999 & 0.991 & 0.090 \\
base & Qwen3.5-35B-A3B & pico-zero-shot & 0 & 0.410 & 0.431 & 0.396 & 0.110 & 0.997 & 0.997 & 0.092 \\
base & Qwen3.5-35B-A3B & zero-shot & 0 & 0.444 & 0.471 & 0.428 & 0.167 & 0.985 & 0.985 & 0.094 \\
\bottomrule
\end{tabular*}
\caption{All complete paper-facing runs (1,752 rows each). Classifier rows are verdict-only, so evidence and ROUGE-L are not applicable (part 1 of 3).}
\label{tab:app-complete-results}
\end{table*}

\begin{table*}[!t]
\centering
\small
\setlength{\tabcolsep}{1.5pt}
\begin{tabular*}{\textwidth}{@{\extracolsep{\fill}}lllrrrrrrrr@{}}
\toprule
Reg. & Model & Prompt & Seed & F1 & Acc. & Bal. & MCC & Valid & Evid. & R-L \\
\midrule
base & Qwen3.5-4B & cot-fewshot & 0 & 0.390 & 0.385 & 0.393 & 0.099 & 0.999 & 0.999 & 0.085 \\
base & Qwen3.5-4B & pico-fewshot & 0 & 0.366 & 0.364 & 0.372 & 0.068 & 0.999 & 0.999 & 0.079 \\
base & Qwen3.5-4B & pico-zero-shot & 0 & 0.351 & 0.354 & 0.365 & 0.064 & 1.000 & 1.000 & 0.088 \\
base & Qwen3.5-4B & zero-shot & 0 & 0.418 & 0.429 & 0.410 & 0.131 & 0.996 & 0.996 & 0.090 \\
base & Qwen3.5-9B & cot-fewshot & 0 & 0.393 & 0.401 & 0.392 & 0.106 & 0.999 & 0.998 & 0.089 \\
base & Qwen3.5-9B & pico-fewshot & 0 & 0.357 & 0.374 & 0.368 & 0.067 & 1.000 & 0.999 & 0.087 \\
base & Qwen3.5-9B & pico-zero-shot & 0 & 0.415 & 0.427 & 0.405 & 0.123 & 1.000 & 1.000 & 0.088 \\
base & Qwen3.5-9B & zero-shot & 0 & 0.431 & 0.459 & 0.422 & 0.152 & 1.000 & 1.000 & 0.094 \\
base & Qwen3.6-27B & cot-fewshot & 0 & 0.430 & 0.442 & 0.419 & 0.143 & 1.000 & 1.000 & 0.088 \\
base & Qwen3.6-27B & pico-fewshot & 0 & 0.408 & 0.409 & 0.400 & 0.110 & 1.000 & 1.000 & 0.086 \\
base & Qwen3.6-27B & pico-zero-shot & 0 & 0.419 & 0.426 & 0.406 & 0.120 & 0.999 & 0.999 & 0.086 \\
base & Qwen3.6-27B & zero-shot & 0 & 0.434 & 0.470 & 0.431 & 0.166 & 1.000 & 1.000 & 0.089 \\
base & Qwen3.6-35B-A3B & cot-fewshot & 0 & 0.422 & 0.458 & 0.409 & 0.132 & 0.999 & 0.985 & 0.091 \\
base & Qwen3.6-35B-A3B & pico-zero-shot & 0 & 0.419 & 0.443 & 0.407 & 0.125 & 1.000 & 0.995 & 0.085 \\
base & Qwen3.6-35B-A3B & zero-shot & 0 & 0.437 & 0.481 & 0.425 & 0.164 & 0.994 & 0.993 & 0.092 \\
rag & Gemma-4-26B-A4B & cot-fewshot & 0 & 0.371 & 0.394 & 0.370 & 0.078 & 1.000 & 1.000 & 0.081 \\
rag & Gemma-4-26B-A4B & pico-fewshot & 0 & 0.390 & 0.408 & 0.382 & 0.094 & 1.000 & 1.000 & 0.081 \\
rag & Gemma-4-26B-A4B & pico-zero-shot & 0 & 0.397 & 0.426 & 0.389 & 0.108 & 0.999 & 1.000 & 0.089 \\
rag & Gemma-4-26B-A4B & zero-shot & 0 & 0.383 & 0.412 & 0.380 & 0.097 & 0.999 & 0.999 & 0.090 \\
rag & Gemma-4-31B & cot-fewshot & 0 & 0.399 & 0.415 & 0.393 & 0.109 & 0.999 & 0.999 & 0.084 \\
rag & Gemma-4-31B & pico-fewshot & 0 & 0.409 & 0.424 & 0.401 & 0.120 & 1.000 & 1.000 & 0.083 \\
rag & Gemma-4-31B & pico-zero-shot & 0 & 0.400 & 0.416 & 0.391 & 0.108 & 1.000 & 1.000 & 0.088 \\
rag & Gemma-4-31B & zero-shot & 0 & 0.396 & 0.412 & 0.389 & 0.108 & 0.992 & 0.992 & 0.090 \\
rag & Gemma-4-E2B & cot-fewshot & 0 & 0.368 & 0.421 & 0.376 & 0.090 & 0.999 & 0.999 & 0.082 \\
rag & Gemma-4-E2B & pico-fewshot & 0 & 0.376 & 0.434 & 0.380 & 0.104 & 0.993 & 0.998 & 0.083 \\
rag & Gemma-4-E2B & pico-zero-shot & 0 & 0.344 & 0.389 & 0.360 & 0.059 & 0.999 & 0.999 & 0.092 \\
rag & Gemma-4-E2B & zero-shot & 0 & 0.295 & 0.376 & 0.344 & 0.035 & 0.999 & 0.999 & 0.095 \\
rag & Gemma-4-E4B & cot-fewshot & 0 & 0.368 & 0.400 & 0.371 & 0.079 & 0.999 & 0.999 & 0.085 \\
rag & Gemma-4-E4B & pico-fewshot & 0 & 0.388 & 0.416 & 0.384 & 0.095 & 0.997 & 1.000 & 0.084 \\
rag & Gemma-4-E4B & pico-zero-shot & 0 & 0.367 & 0.411 & 0.382 & 0.096 & 0.999 & 1.000 & 0.090 \\
rag & Gemma-4-E4B & zero-shot & 0 & 0.328 & 0.380 & 0.362 & 0.068 & 1.000 & 1.000 & 0.093 \\
rag & Qwen3.5-27B & cot-fewshot & 0 & 0.407 & 0.434 & 0.399 & 0.119 & 0.999 & 0.999 & 0.090 \\
rag & Qwen3.5-27B & pico-fewshot & 0 & 0.421 & 0.448 & 0.408 & 0.138 & 0.995 & 0.999 & 0.092 \\
rag & Qwen3.5-27B & pico-zero-shot & 0 & 0.399 & 0.422 & 0.386 & 0.113 & 0.975 & 0.975 & 0.091 \\
rag & Qwen3.5-27B & zero-shot & 0 & 0.393 & 0.416 & 0.383 & 0.104 & 0.986 & 0.986 & 0.094 \\
rag & Qwen3.5-2B & cot-fewshot & 0 & 0.368 & 0.437 & 0.370 & 0.085 & 0.987 & 0.987 & 0.087 \\
rag & Qwen3.5-2B & pico-fewshot & 0 & 0.362 & 0.403 & 0.351 & 0.063 & 0.966 & 0.966 & 0.082 \\
rag & Qwen3.5-2B & pico-zero-shot & 0 & 0.385 & 0.436 & 0.384 & 0.092 & 0.998 & 0.998 & 0.095 \\
\bottomrule
\end{tabular*}
\caption{All complete paper-facing runs (1,752 rows each). Classifier rows are verdict-only, so evidence and ROUGE-L are not applicable (part 2 of 3).}
\label{tab:app-complete-results-part-2}
\end{table*}

\begin{table*}[!t]
\centering
\small
\setlength{\tabcolsep}{1.5pt}
\begin{tabular*}{\textwidth}{@{\extracolsep{\fill}}lllrrrrrrrr@{}}
\toprule
Reg. & Model & Prompt & Seed & F1 & Acc. & Bal. & MCC & Valid & Evid. & R-L \\
\midrule
rag & Qwen3.5-2B & zero-shot & 0 & 0.320 & 0.423 & 0.356 & 0.058 & 1.000 & 1.000 & 0.097 \\
rag & Qwen3.5-35B-A3B & cot-fewshot & 0 & 0.390 & 0.431 & 0.380 & 0.110 & 0.976 & 0.971 & 0.092 \\
rag & Qwen3.5-35B-A3B & pico-zero-shot & 0 & 0.400 & 0.438 & 0.389 & 0.123 & 0.981 & 0.981 & 0.094 \\
rag & Qwen3.5-35B-A3B & zero-shot & 0 & 0.393 & 0.425 & 0.387 & 0.110 & 0.993 & 0.993 & 0.097 \\
rag & Qwen3.5-4B & cot-fewshot & 0 & 0.365 & 0.414 & 0.362 & 0.067 & 1.000 & 1.000 & 0.083 \\
rag & Qwen3.5-4B & pico-fewshot & 0 & 0.382 & 0.421 & 0.375 & 0.083 & 1.000 & 1.000 & 0.085 \\
rag & Qwen3.5-4B & pico-zero-shot & 0 & 0.380 & 0.421 & 0.374 & 0.086 & 0.999 & 0.987 & 0.089 \\
rag & Qwen3.5-4B & zero-shot & 0 & 0.355 & 0.402 & 0.358 & 0.063 & 1.000 & 0.999 & 0.091 \\
rag & Qwen3.5-9B & cot-fewshot & 0 & 0.401 & 0.434 & 0.393 & 0.112 & 0.999 & 1.000 & 0.089 \\
rag & Qwen3.5-9B & pico-fewshot & 0 & 0.414 & 0.445 & 0.403 & 0.129 & 0.998 & 1.000 & 0.087 \\
rag & Qwen3.5-9B & pico-zero-shot & 0 & 0.413 & 0.460 & 0.404 & 0.136 & 0.999 & 0.999 & 0.089 \\
rag & Qwen3.5-9B & zero-shot & 0 & 0.431 & 0.461 & 0.421 & 0.155 & 0.999 & 1.000 & 0.091 \\
rag & Qwen3.6-27B & cot-fewshot & 0 & 0.390 & 0.420 & 0.379 & 0.090 & 1.000 & 1.000 & 0.088 \\
rag & Qwen3.6-27B & pico-fewshot & 0 & 0.391 & 0.416 & 0.379 & 0.088 & 0.999 & 1.000 & 0.088 \\
rag & Qwen3.6-27B & pico-zero-shot & 0 & 0.411 & 0.440 & 0.400 & 0.121 & 1.000 & 1.000 & 0.090 \\
rag & Qwen3.6-27B & zero-shot & 0 & 0.405 & 0.432 & 0.394 & 0.111 & 1.000 & 1.000 & 0.093 \\
rag & Qwen3.6-35B-A3B & cot-fewshot & 0 & 0.391 & 0.431 & 0.386 & 0.103 & 0.999 & 0.987 & 0.089 \\
rag & Qwen3.6-35B-A3B & pico-fewshot & 0 & 0.422 & 0.455 & 0.411 & 0.138 & 0.999 & 0.997 & 0.090 \\
rag & Qwen3.6-35B-A3B & pico-zero-shot & 0 & 0.405 & 0.453 & 0.400 & 0.135 & 0.995 & 0.994 & 0.094 \\
rag & Qwen3.6-35B-A3B & zero-shot & 0 & 0.399 & 0.434 & 0.390 & 0.125 & 0.977 & 0.976 & 0.095 \\
FT & Gemma-4-31B & cot-fewshot & 0 & 0.480 & 0.538 & 0.473 & 0.243 & 1.000 & 1.000 & 0.053 \\
FT & Gemma-4-31B & pico-fewshot & 0 & 0.483 & 0.538 & 0.476 & 0.247 & 1.000 & 1.000 & 0.053 \\
FT & Gemma-4-E4B & cot-fewshot & 0 & 0.476 & 0.576 & 0.491 & 0.300 & 0.990 & 0.990 & 0.076 \\
FT & Qwen3.5-35B-A3B & cot-fewshot & 0 & 0.481 & 0.576 & 0.485 & 0.294 & 0.998 & 0.986 & 0.081 \\
FT & Qwen3.5-4B & cot-fewshot & 0 & 0.435 & 0.456 & 0.430 & 0.164 & 0.999 & 0.994 & 0.072 \\
FT & Qwen3.5-4B & pico-fewshot & 0 & 0.396 & 0.395 & 0.402 & 0.117 & 1.000 & 0.999 & 0.070 \\
FT & Qwen3.5-4B & zero-shot & 0 & 0.431 & 0.462 & 0.419 & 0.153 & 1.000 & 0.981 & 0.084 \\
FT & Qwen3.6-27B & cot-fewshot & 0 & 0.381 & 0.385 & 0.395 & 0.113 & 1.000 & 1.000 & 0.072 \\
FT & Qwen3.6-27B & pico-fewshot & 0 & 0.386 & 0.408 & 0.386 & 0.104 & 1.000 & 1.000 & 0.075 \\
FT & Qwen3.6-27B & zero-shot & 0 & 0.478 & 0.577 & 0.487 & 0.295 & 0.999 & 0.999 & 0.086 \\
FT & Qwen3.6-35B-A3B & cot-fewshot & 0 & 0.451 & 0.494 & 0.441 & 0.196 & 1.000 & 0.999 & 0.080 \\
FT & Qwen3.6-35B-A3B & pico-zero-shot & 0 & 0.464 & 0.553 & 0.465 & 0.253 & 1.000 & 1.000 & 0.086 \\
FT & Qwen3.6-35B-A3B & zero-shot & 0 & 0.474 & 0.581 & 0.483 & 0.300 & 1.000 & 1.000 & 0.089 \\
\bottomrule
\end{tabular*}
\caption{All complete paper-facing runs (1,752 rows each). Classifier rows are verdict-only, so evidence and ROUGE-L are not applicable (part 3 of 3).}
\label{tab:app-complete-results-part-3}
\end{table*}

\begin{table*}[!t]
\centering
\small
\setlength{\tabcolsep}{2pt}
\begin{tabularx}{\textwidth}{@{}l>{\raggedright\arraybackslash}p{0.15\textwidth}rX>{\raggedright\arraybackslash}p{0.14\textwidth}>{\raggedright\arraybackslash}p{0.10\textwidth}rr@{}}
\toprule
Reg. & Model & Runs & Prompts & Reason & Rows & Valid & Evid. \\
\midrule
base & Gemini-3.1-Flash-Lite & 4 & cot-fewshot, pico-fewshot, pico-zero-shot, zero-shot & low valid label coverage & 1752--1752/1752 & 0.850--0.916 & 0.850--0.916 \\
base & Gemma-4-E4B & 3 & cot-fewshot, pico-zero-shot, zero-shot & incomplete row count & 1751--1751/1752 & 0.998--0.999 & 0.998--0.999 \\
base & Qwen3.6-35B-A3B & 1 & pico-fewshot & low nonempty evidence coverage & 1752--1752/1752 & 0.998--0.998 & 0.934--0.934 \\
baseline & oracle-retrieval & 1 & claim-only & planned not run & 0--0/1752 & 0.000--0.000 & 0.000--0.000 \\
baseline & pubmed-rank-only-rag & 1 & claim-only & planned not run & 0--0/1752 & 0.000--0.000 & 0.000--0.000 \\
FT & Gemma-4-26B-A4B & 4 & cot-fewshot, pico-fewshot, pico-zero-shot, zero-shot & template copy & 1752--1752/1752 & 0.001--0.669 & 0.961--0.999 \\
FT & Gemma-4-31B & 2 & pico-zero-shot, zero-shot & template copy & 1752--1752/1752 & 0.000--0.001 & 1.000--1.000 \\
FT & Gemma-4-E2B & 4 & cot-fewshot, pico-fewshot, pico-zero-shot, zero-shot & template copy & 1752--1752/1752 & 0.078--0.772 & 0.961--0.990 \\
FT & Gemma-4-E4B & 6 & cot-fewshot, pico-fewshot, pico-zero-shot, zero-shot & incomplete row count & 205--762/1752 & 0.117--0.435 & 0.117--0.435 \\
FT & Gemma-4-E4B & 3 & pico-fewshot, pico-zero-shot, zero-shot & template copy & 1752--1752/1752 & 0.000--0.982 & 0.993--1.000 \\
FT & Qwen3.5-2B & 4 & cot-fewshot, pico-fewshot, pico-zero-shot, zero-shot & template copy & 1752--1752/1752 & 0.000--0.000 & 0.627--0.760 \\
FT & Qwen3.5-35B-A3B & 2 & pico-zero-shot, zero-shot & low nonempty evidence coverage & 1752--1752/1752 & 0.999--1.000 & 0.909--0.935 \\
FT & Qwen3.5-35B-A3B & 1 & pico-fewshot & template copy & 1752--1752/1752 & 0.913--0.913 & 0.981--0.981 \\
FT & Qwen3.5-4B & 1 & pico-zero-shot & low nonempty evidence coverage & 1752--1752/1752 & 0.999--0.999 & 0.616--0.616 \\
FT & Qwen3.5-9B & 4 & cot-fewshot, pico-fewshot, pico-zero-shot, zero-shot & low valid label coverage & 1752--1752/1752 & 0.000--0.000 & 0.000--0.000 \\
FT & Qwen3.6-27B & 1 & pico-zero-shot & template copy & 1752--1752/1752 & 0.999--0.999 & 1.000--1.000 \\
FT & Qwen3.6-35B-A3B & 1 & pico-fewshot & template copy & 1752--1752/1752 & 0.995--0.995 & 0.999--0.999 \\
FT & Qwen3.6-27B & 3 & zero-shot & incomplete row count & 1216--1293/1752 & 0.694--0.738 & 0.694--0.738 \\
FT & Qwen3.6-35B-A3B & 3 & zero-shot & incomplete row count & 984--1198/1752 & 0.562--0.684 & 0.562--0.684 \\
rag & Gemini-3.1-Flash-Lite & 4 & cot-fewshot, pico-fewshot, pico-zero-shot, zero-shot & incomplete row count & 1714--1714/1752 & 0.978--0.978 & 0.978--0.978 \\
rag & Qwen3.5-35B-A3B & 1 & pico-fewshot & low nonempty evidence coverage & 1752--1752/1752 & 0.954--0.954 & 0.946--0.946 \\
\bottomrule
\end{tabularx}
\caption{Quality-gated outputs retained in the manifest but excluded from headline claims.}
\label{tab:app-failures}
\end{table*}

The complete-results table should be read together with the failure table. Several planned model--strategy combinations are retained as missing or incomplete because they were part of the experiment design but do not provide usable aligned outputs. This avoids a common reporting bias in generative-model evaluation: failed generations, invalid schemas, and low-coverage outputs are easy to omit, but omitting them makes evidence generation appear more robust than it is. The manifest policy therefore separates \emph{experiment attempted or planned} from \emph{experiment usable for headline comparison}.

\begin{table}[t]
\centering
\small
\begin{tabular}{llr}
\toprule
Phase/regime & Status & Count \\
\midrule
Current base & Complete & 44 \\
Current base & Incomplete & 8 \\
Current RAG & Complete & 43 \\
Current RAG & Incomplete & 5 \\
Current fine-tuned & Complete & 13 \\
Current fine-tuned & Incomplete & 27 \\
Classifier baselines v2 & Complete & 18 \\
Classifier baselines & Complete & 9 \\
Label-only LLM & Complete & 8 \\
Simple baselines & Complete & 4 \\
Reduced fine-tuned reruns & Incomplete & 12 \\
\bottomrule
\end{tabular}
\caption{Manifest bookkeeping in the current paper-results export. Missing planned runs are retained explicitly; incomplete runs retain failure reasons and coverage instead of headline metrics.}
\label{tab:app-manifest}
\end{table}

The current grids now contain 44 complete base runs, 43 complete RAG runs, and 13 complete fine-tuned runs. Remaining attempted outputs are retained as incomplete when they fail row-count, label-coverage, evidence-coverage, or template-copy quality gates. Classifier baselines and label-only LLM diagnostics are compact targeted grids and therefore have higher completion rates. Reduced fine-tuned rerun rows are kept incomplete when they do not cover the full aligned test set.

\section{Full Dataset Details}

CARE-XAI combines five source datasets into a shared three-way label space. This is useful for a unified biomedical verification study, but it also creates a heterogeneous evidence landscape. PubMedQA and SciFact are closest to scientific abstract verification; HealthVer and PUBHEALTH contain broader health claims; HealthFC is smaller but useful for checking health fact-checking generalization. Table~\ref{tab:app-dataset} gives the exact split composition used throughout the paper.

\begin{table*}[t]
\centering
\small
\begin{tabular}{lrrrrrrrrr}
\toprule
Split & Rows & PubMedQA & SciFact & HealthVer & PUBHEALTH & HealthFC & Supp. & Contr. & Unaddr. \\
\midrule
Train & 14,254 & 782 & 761 & 4,238 & 7,866 & 607 & 6,840 & 4,001 & 3,413 \\
Validation & 1,797 & 113 & 106 & 544 & 964 & 70 & 841 & 565 & 391 \\
Test & 1,752 & 105 & 90 & 510 & 974 & 73 & 851 & 495 & 406 \\
\bottomrule
\end{tabular}
\caption{Full CARE-XAI split composition by source and label.}
\label{tab:app-dataset}
\end{table*}

\section{Retrieval Diagnostics}

The frozen retrieval cache makes retrieval analysis independent of the downstream answer model. Every RAG generation consumes the same cached PubMed candidates and reranked contexts, so differences between base/RAG/fine-tuned outputs are not caused by repeated live PubMed calls. Table~\ref{tab:app-retrieval} summarizes global cache quality, while Table~\ref{tab:app-retrieval-source} shows that retrieval quality is sharply source-dependent.

\begin{table}[t]
\centering
\small
\setlength{\tabcolsep}{2pt}
\begin{tabular}{lrrrrrr}
\toprule
Rows & Query & MRR & R@1 & R@5 & R@10 & nDCG \\
\midrule
1,752 & 0.871 & 0.212 & 0.205 & 0.220 & 0.224 & 0.214 \\
\bottomrule
\end{tabular}
\caption{Frozen PubMed retrieval-cache diagnostics. Relevance metrics are computed on 477 rows with known source PMIDs.}
\label{tab:app-retrieval}
\end{table}

\begin{table*}[t]
\centering
\small
\begin{tabular}{lrrrrrrrr}
\toprule
Source & Rows & Query & Docs & Rel. rows & MRR & R@1 & R@10 & Hit@10 \\
\midrule
HealthFC & 73 & 0.973 & 138.1 & 73 & 0.000 & 0.000 & 0.000 & 0.000 \\
HealthVer & 510 & 0.990 & 176.1 & 188 & 0.035 & 0.021 & 0.064 & 0.064 \\
PUBHEALTH & 974 & 0.777 & 90.7 & 21 & 0.029 & 0.000 & 0.048 & 0.048 \\
PubMedQA & 105 & 1.000 & 128.9 & 105 & 0.895 & 0.895 & 0.895 & 0.895 \\
SciFact & 90 & 0.978 & 116.1 & 90 & 0.000 & 0.000 & 0.000 & 0.000 \\
\bottomrule
\end{tabular}
\caption{Source-stratified retrieval diagnostics. PubMedQA is the only source where source PMID retrieval is consistently high.}
\label{tab:app-retrieval-source}
\end{table*}

Figure~\ref{fig:retrieval-diagnostics-main} in the main paper visualizes the important pattern: retrieval can be operationally successful without being decision-useful. A query may return PubMed abstracts and BGE may assign high topical relevance, but the retrieved abstract can still fail to address the exact claim. This is especially visible for public-health claims whose correct evidence may be a news report, a policy document, or a fact-checking explanation rather than a biomedical abstract.

\section{Evidence Utility Diagnostics}

\subsection{Formal Bio-GRACE Supplement}

Bio-GRACE evaluates retrieval as a counterfactual intervention on a verifier. For example \(i\), let \(p_i^C(y_i)\), \(p_i^R(y_i)\), and \(p_i^G(y_i)\) denote the verifier probability assigned to the gold label under claim-only, retrieved-context, and gold-evidence inputs. The oracle evidence gain and retrieved evidence gain are:
\begin{align}
U_i^\star &= p_i^G(y_i)-p_i^C(y_i),\\
U_i^R &= p_i^R(y_i)-p_i^C(y_i).
\end{align}
Bio-GRACE focuses on evidence-sensitive examples \(\mathcal{H}=\{i:U_i^\star>0\}\), because these are cases where reference evidence actually improves the verifier. Utility recovery is:
\begin{equation}
\mathrm{UR}=\frac{1}{|\mathcal{H}|}\sum_{i\in\mathcal{H}}\mathrm{clip}\left(\frac{U_i^R}{U_i^\star+\epsilon},-1,1\right).
\end{equation}
The clipping makes the score bounded in \([-1,1]\). A value near \(1\) means retrieval recovers most of the reference-evidence benefit; a value near \(0\) means retrieval adds little; a negative value means retrieval moves probability away from the true label on evidence-sensitive examples.

\paragraph{Boundedness.}
For every \(i\in\mathcal{H}\), the clipped term lies in \([-1,1]\), hence the arithmetic mean also lies in \([-1,1]\). This matters because \(U_i^\star\) can be very small for some examples; without clipping, a small denominator can make a single example dominate the aggregate score.

\paragraph{Sign interpretation.}
Since \(U_i^\star>0\) on \(\mathcal{H}\), the sign of the unclipped ratio is the sign of \(U_i^R\). Positive UR therefore indicates that retrieval usually increases the true-label probability relative to the claim-only setting. Negative UR indicates retrieval usually decreases it. This is the core distinction between topical retrieval and decision-useful retrieval.

\paragraph{Relation to rescue and distraction.}
Bio-GRACE complements two discrete paired rates. A rescue occurs when the claim-only prediction is wrong and the retrieved-context prediction is correct. A distraction occurs when the claim-only prediction is correct and the retrieved-context prediction is wrong. Net retrieval improvement (NRI) is the rescue rate minus the distraction rate. UR is probability-sensitive, while NRI is decision-boundary-sensitive; both are useful because retrieval can improve confidence without changing the argmax label, or cross the decision boundary in either direction.

\paragraph{Why gold evidence is allowed.}
Gold/reference evidence is not used as a deployment input. It is used only to define whether the example is evidence-sensitive and to normalize how much benefit retrieval recovers. This makes Bio-GRACE an evaluation diagnostic: it asks whether a retrieval method approximates the decision benefit of trusted evidence, not whether the system has access to gold evidence at test time.

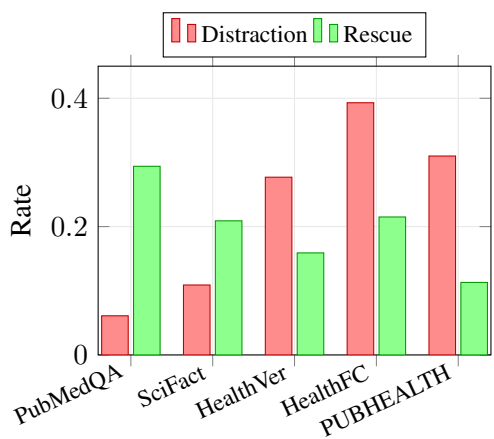
\begin{figure}[t]
\centering
\begin{tikzpicture}
\begin{axis}[
  ybar,
  width=0.88\columnwidth,
  height=5.4cm,
  ylabel={Rate},
  symbolic x coords={PubMedQA,SciFact,HealthVer,HealthFC,PUBHEALTH},
  xtick=data,
  x tick label style={rotate=25,anchor=east,font=\small},
  ymin=0,
  ymax=0.45,
  bar width=10pt,
  legend style={at={(0.5,1.03)},anchor=south,legend columns=2,font=\small},
  grid=major,
  grid style={draw=gray!18},
]
\addplot+[fill=red!45,draw=red!75!black] coordinates {(PubMedQA,0.061) (SciFact,0.109) (HealthVer,0.277) (HealthFC,0.393) (PUBHEALTH,0.310)};
\addplot+[fill=green!45,draw=green!60!black] coordinates {(PubMedQA,0.294) (SciFact,0.209) (HealthVer,0.159) (HealthFC,0.215) (PUBHEALTH,0.113)};
\legend{Distraction,Rescue}
\end{axis}
\end{tikzpicture}
\caption{Retrieval distraction and rescue rates by source. Public-health sources show higher distraction than rescue, explaining why always-on PubMed retrieval can degrade aggregate behavior.}
\label{fig:app-biograce-distraction-rescue}
\end{figure}

Table~\ref{tab:app-biograce-source} expands Bio-GRACE by source, and Figure~\ref{fig:app-biograce-distraction-rescue} shows the paired distraction and rescue rates behind the utility score. Table~\ref{tab:app-biograce-validation} provides probability-level validation checks, while Table~\ref{tab:app-bge} relates retrieval confidence to downstream utility.

\begin{table*}[t]
\centering
\small
\begin{tabular}{lrrrrrrr}
\toprule
Source & $N$ & Unique & UR & CW-UR & NRI & Distraction & Rescue \\
\midrule
PubMedQA & 945 & 105 & 0.676 & 0.763 & 0.233 & 0.061 & 0.294 \\
SciFact & 810 & 90 & 0.435 & 0.536 & 0.100 & 0.109 & 0.209 \\
HealthVer & 4,590 & 510 & -0.030 & 0.089 & -0.118 & 0.277 & 0.159 \\
HealthFC & 657 & 73 & -0.222 & 0.029 & -0.178 & 0.393 & 0.215 \\
PUBHEALTH & 8,766 & 974 & -0.378 & -0.018 & -0.197 & 0.310 & 0.113 \\
\bottomrule
\end{tabular}
\caption{Source-stratified Bio-GRACE diagnostics. Counts are model-seed rows from three classifiers and three seeds; unique rows divide $N$ by 9.}
\label{tab:app-biograce-source}
\end{table*}

\begin{table}[t]
\centering
\small
\begin{tabularx}{\columnwidth}{@{}Xrr@{}}
\toprule
Check & Value & $N$ \\
\midrule
Gold positive-utility rate & 0.674 & 15,768 \\
Retrieved positive-utility rate & 0.373 & 15,768 \\
Mean oracle log-utility & 0.198 & 15,768 \\
Mean retrieved log-utility & -0.564 & 15,768 \\
UR vs retrieved F1-lift Spearman & 0.183 & 9 \\
Source UR vs retrieved acc.-lift Spearman & 1.000 & 5 \\
\bottomrule
\end{tabularx}
\caption{Lightweight Bio-GRACE validation checks from stored classifier probability artifacts.}
\label{tab:app-biograce-validation}
\end{table}

\begin{table*}[t]
\centering
\small
\begin{tabular}{lrrrrr}
\toprule
Source & Top BGE & Docs & UR & Ret. acc. lift & Hit@10 \\
\midrule
HealthFC & 5.094 & 138.1 & -0.222 & -0.178 & 0.000 \\
HealthVer & 3.294 & 176.1 & -0.030 & -0.118 & 0.064 \\
PUBHEALTH & -0.385 & 90.7 & -0.378 & -0.197 & 0.048 \\
PubMedQA & 8.469 & 128.9 & 0.676 & 0.233 & 0.895 \\
SciFact & 3.793 & 116.1 & 0.435 & 0.100 & 0.000 \\
\midrule
\multicolumn{3}{l}{Pearson(top BGE, UR)} & \multicolumn{3}{r}{0.762} \\
\multicolumn{3}{l}{Pearson(top BGE, retrieved accuracy lift)} & \multicolumn{3}{r}{0.734} \\
\bottomrule
\end{tabular}
\caption{Exploratory retrieval-score diagnostics. BGE score aligns descriptively with retrieval utility across five sources, while document count and PubMed result count were weak signals.}
\label{tab:app-bge}
\end{table*}

Figure~\ref{fig:biograce-source-main} in the main paper and Figure~\ref{fig:app-biograce-distraction-rescue} visualize the same source-level mechanism from two perspectives. UR shows how much retrieved context recovers reference-evidence benefit; distraction/rescue rates show how often retrieval flips a paired prediction in the wrong or right direction. The two views agree qualitatively: PubMed-aligned sources benefit most, while public-health sources are more vulnerable to retrieval-induced distraction.

\section{Evidence-Conditioning and Verdict-Only Diagnostics}

\begin{figure}[t]
\centering
\includegraphics[width=\columnwidth]{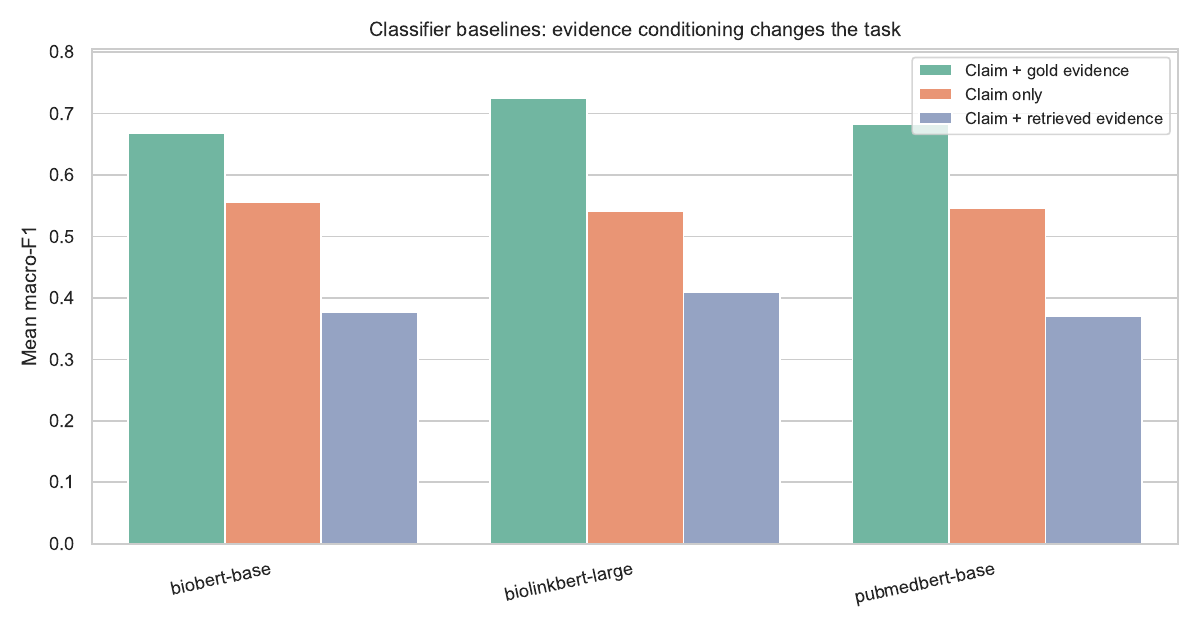}
\caption{Classifier macro-F1 under claim-only, retrieved-evidence, and gold-evidence protocols. Gold evidence is an oracle diagnostic and should not be interpreted as a deployable baseline.}
\label{fig:app-classifier-protocols}
\end{figure}

\begin{figure}[t]
\centering
\includegraphics[width=\columnwidth]{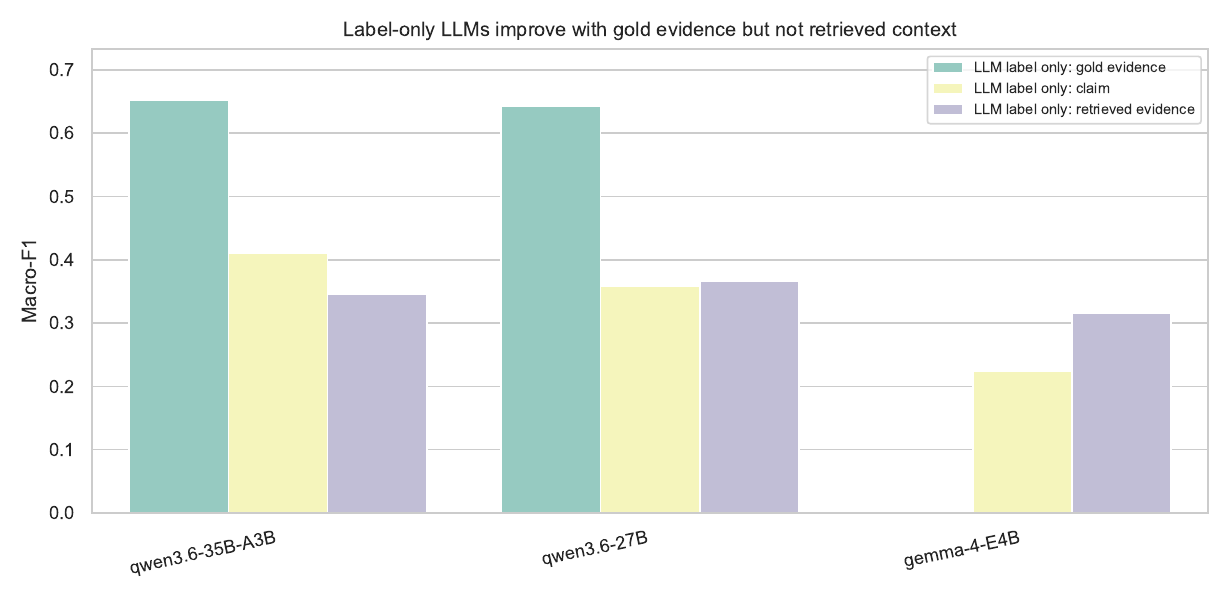}
\caption{Label-only LLM protocol comparison. Removing evidence generation isolates whether LLMs can use supplied evidence for verdict prediction.}
\label{fig:app-llm-label-only}
\end{figure}

The main paper distinguishes evidence-generating systems from verdict-only probes. This section expands that distinction with additional plots. Figure~\ref{fig:app-classifier-protocols} compares classifier input protocols, while Figure~\ref{fig:app-llm-label-only} isolates label prediction in LLMs. Together, these evidence-conditioned probes ask whether reference evidence is decision-useful and whether retrieved evidence closes the gap to reference evidence. They are not evidence-generating verifiers, but they help identify which component is failing.

The diagnostic picture is consistent with Bio-GRACE. Gold evidence improves verdict prediction substantially, so the task is not label-noise dominated. Retrieved PubMed evidence does not reliably recover this benefit, so the main bottleneck is evidence retrieval and source matching rather than a complete inability to use evidence.

\section{Leakage Sensitivity}

\begin{figure}[t]
\centering
\includegraphics[width=\columnwidth]{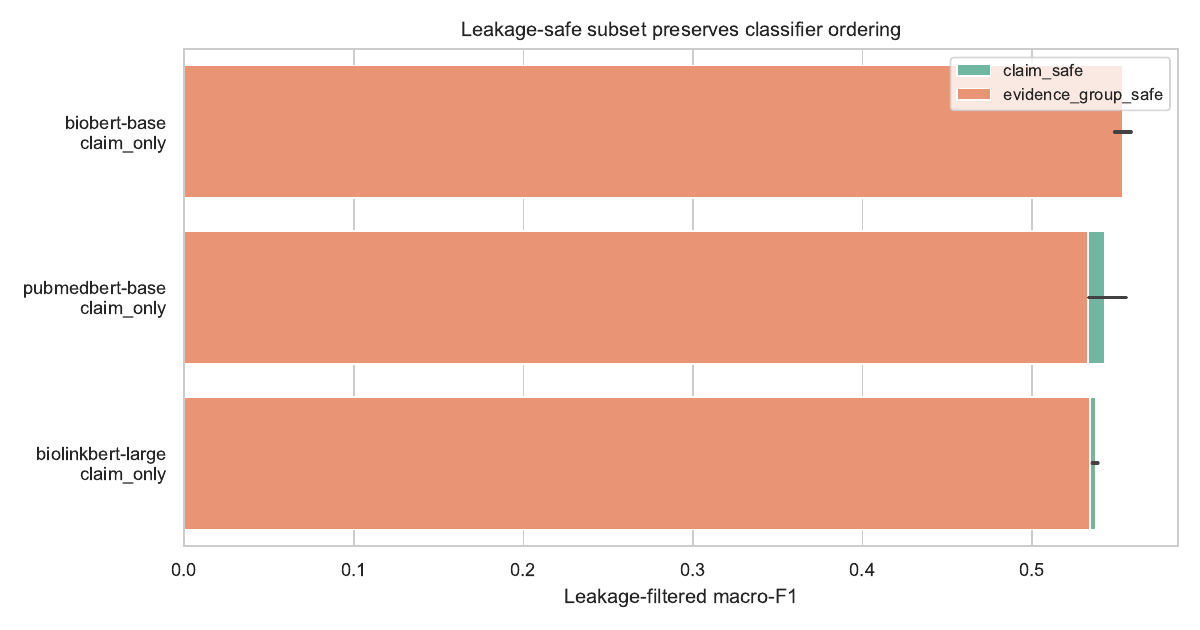}
\caption{Full-test versus leakage-filtered performance. The qualitative ordering remains stable after evidence-group-safe filtering.}
\label{fig:app-leakage-filtered}
\end{figure}

The leakage audit checks whether conclusions survive when evidence-overlap rows are removed. Table~\ref{tab:app-leakage} reports the regime-level changes and Figure~\ref{fig:app-leakage-filtered} visualizes the retained-system ordering. CARE-XAI is useful because it provides a unified multi-source benchmark, but it is not leakage-free. We therefore report leakage-safe summaries as sensitivity analyses and avoid claiming that official-split results alone are sufficient evidence of generalization.

\begin{table}[t]
\centering
\small
\begin{tabular}{lccc}
\toprule
Regime & Full F1 & Safe F1 & $\Delta$ \\
\midrule
Classifier & 0.5472 & 0.5294 & +0.0179 \\
Base LLM & 0.3940 & 0.3564 & +0.0375 \\
RAG LLM & 0.3855 & 0.3607 & +0.0247 \\
Fine-tuned LLM & 0.4475 & 0.4413 & +0.0062 \\
\bottomrule
\end{tabular}
\caption{Leakage impact by regime. Safe F1 removes rows in the evidence-group-safe filter ($N=591$ removed).}
\label{tab:app-leakage}
\end{table}

The leakage audit should be interpreted as sensitivity analysis rather than as a replacement benchmark. The main conclusion is stable: retrieval remains mixed, fine-tuning remains more reliable than RAG for evidence-generating LLMs, and classifiers remain strongest for verdict-only prediction.

\section{Per-Source Verdict Performance}

\begin{figure}[t]
\centering
\includegraphics[width=\columnwidth]{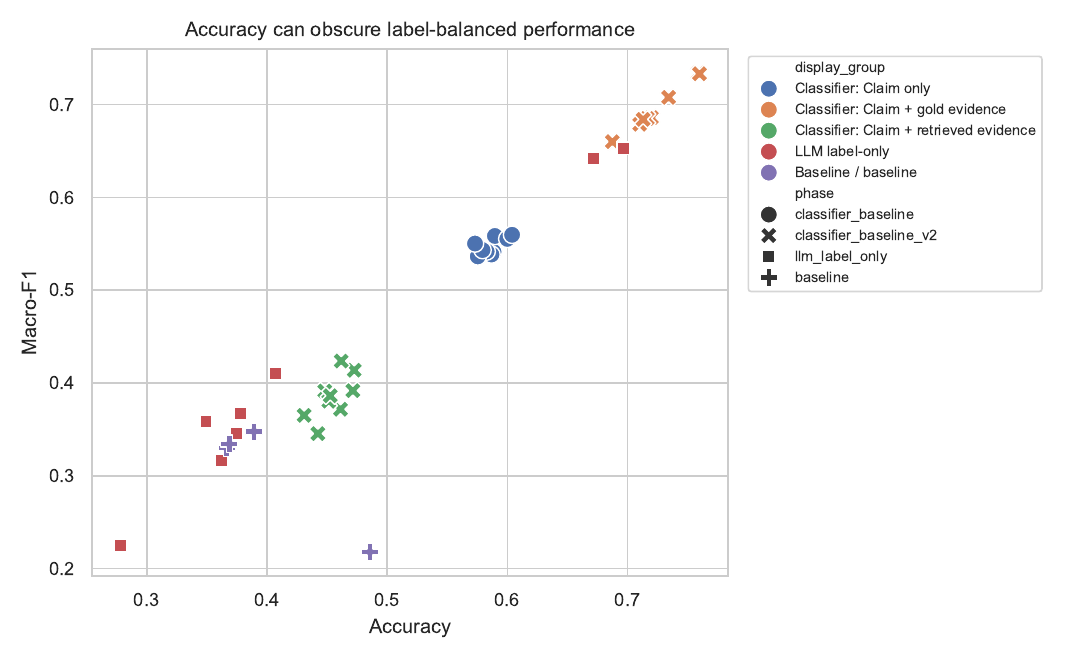}
\caption{Macro-F1 versus accuracy for paper-facing runs. Macro-F1 is emphasized because class imbalance makes accuracy insufficient for evidence verification.}
\label{fig:app-macro-accuracy}
\end{figure}

The pooled verdict tables are useful for a high-level comparison, but source-specific behavior is the more informative diagnostic for retrieval. Table~\ref{tab:app-per-source} and Figure~\ref{fig:app-macro-accuracy} show that systems can have similar aggregate scores while behaving differently across source types.

\begin{table*}[t]
\centering
\small
\begin{tabular}{llrrrrr}
\toprule
Regime & Model & PUBHEALTH & HealthVer & PubMedQA & SciFact & HealthFC \\
\midrule
Classifier & BioBERT-base & 0.513 & 0.542 & 0.353 & 0.423 & 0.445 \\
Base LLM & Qwen3.5-27B & 0.431 & 0.435 & 0.425 & 0.515 & 0.498 \\
RAG LLM & Qwen3.5-9B & 0.356 & 0.450 & 0.524 & 0.615 & 0.437 \\
Fine-tuned LLM & Gemma-4-31B & 0.466 & 0.473 & 0.427 & 0.490 & 0.560 \\
\bottomrule
\end{tabular}
\caption{Per-source macro-F1 for the best model in each regime. RAG is comparatively strong on PubMedQA and SciFact but weaker on PUBHEALTH.}
\label{tab:app-per-source}
\end{table*}

The per-source table is central for interpreting the pooled results. PUBHEALTH dominates the test set by size, so pooled accuracy can obscure source-specific retrieval behavior. PubMedQA and SciFact are closer to PubMed abstract retrieval; HealthFC and PUBHEALTH often require broader public-health, journalistic, or guideline-style context.

\section{Qualitative Retrieval Cases}

Table~\ref{tab:app-qualitative} contrasts representative rescue and distraction cases. The examples are descriptive illustrations of the paired behavior quantified by Bio-GRACE, not evidence for aggregate performance.

\begin{table*}[t]
\centering
\small
\begin{tabular}{llll}
\toprule
Pattern & Source & Claim sketch & Base $\rightarrow$ Retrieved \\
\midrule
Rescue & PubMedQA & ACE level and severe hypoglycaemia in type 1 diabetes & Unaddressed $\rightarrow$ Supported \\
Rescue & PubMedQA & Low serum chloride as cardiovascular mortality risk factor & Unaddressed $\rightarrow$ Supported \\
Distract & PUBHEALTH & Congressional train crash attributed to a ``Deep State'' plot & Contradicted $\rightarrow$ Unaddressed \\
Distract & HealthVer & Hypertension, immune weakness, and COVID-19 risk & Unaddressed $\rightarrow$ Supported \\
\bottomrule
\end{tabular}
\caption{Qualitative routing examples from paired Qwen3.6-27B label-only outputs. Retrieved evidence can rescue source-aligned biomedical claims but distract on broader public-health misinformation claims.}
\label{tab:app-qualitative}
\end{table*}

These examples are not used as proof of general behavior; they illustrate the mechanism detected by Bio-GRACE. Retrieval helps when the retrieved PubMed abstract addresses the same biomedical evidence need. It distracts when the claim requires non-PubMed context, source interpretation, or misinformation-specific context.

\section{NLI Diagnostics}

\begin{figure}[t]
\centering
\includegraphics[width=\columnwidth]{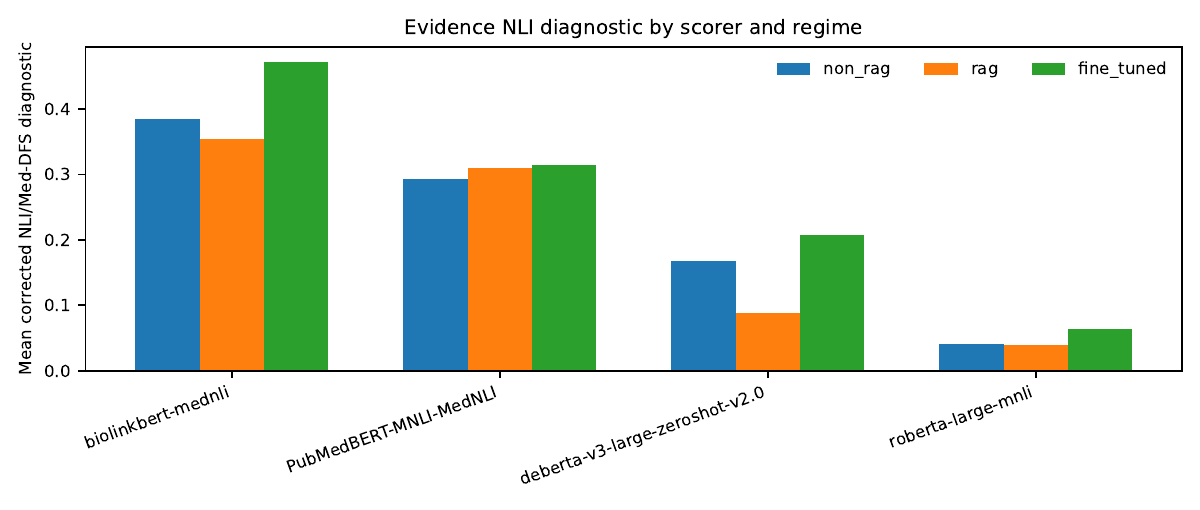}
\caption{NLI entailment behavior by scorer and regime. Biomedical NLI scorers are more permissive than the general-domain RoBERTa-MNLI scorer.}
\label{fig:app-nli-model-regime}
\end{figure}

\begin{figure}[t]
\centering
\includegraphics[width=\columnwidth]{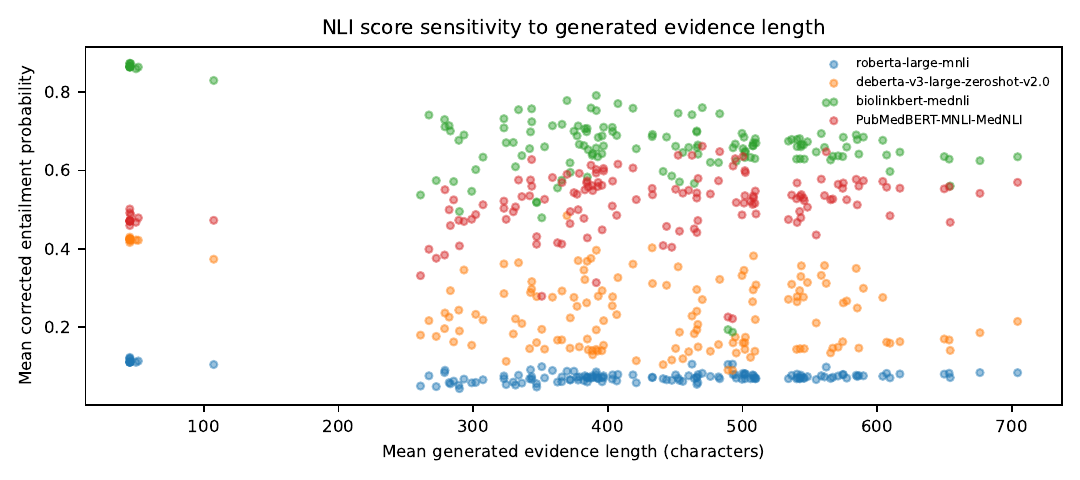}
\caption{NLI length sensitivity. Evidence length and template-like outputs can affect entailment rates, reinforcing that NLI should be interpreted as a diagnostic rather than a final faithfulness metric.}
\label{fig:app-nli-length}
\end{figure}

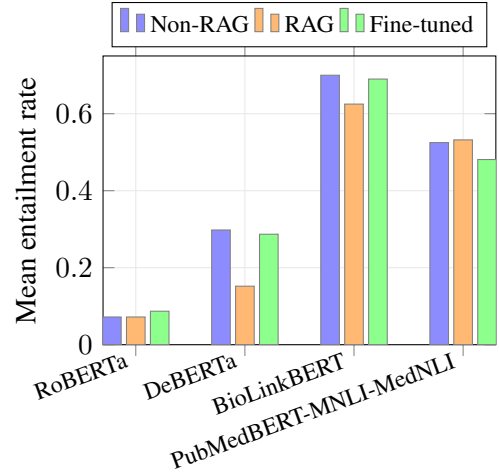
\begin{figure}[t]
\centering
\begin{tikzpicture}
\begin{axis}[
  ybar,
  width=0.88\columnwidth,
  height=5.4cm,
  ylabel={Mean entailment rate},
  symbolic x coords={RoBERTa,DeBERTa,BioLinkBERT,PubMedBERT-MNLI-MedNLI},
  xtick=data,
  x tick label style={rotate=20,anchor=east,font=\small},
  ymin=0,
  ymax=0.75,
  bar width=7pt,
  legend style={at={(0.5,1.03)},anchor=south,legend columns=3,font=\small},
  grid=major,
  grid style={draw=gray!18},
]
\addplot+[fill=blue!45,draw=black!50] coordinates {(RoBERTa,0.072) (DeBERTa,0.298) (BioLinkBERT,0.700) (PubMedBERT-MNLI-MedNLI,0.525)};
\addplot+[fill=orange!55,draw=black!50] coordinates {(RoBERTa,0.072) (DeBERTa,0.152) (BioLinkBERT,0.625) (PubMedBERT-MNLI-MedNLI,0.532)};
\addplot+[fill=green!45,draw=black!50] coordinates {(RoBERTa,0.087) (DeBERTa,0.287) (BioLinkBERT,0.690) (PubMedBERT-MNLI-MedNLI,0.481)};
\legend{Non-RAG,RAG,Fine-tuned}
\end{axis}
\end{tikzpicture}
\caption{Mean entailment rate by NLI scorer and generation regime. The spread across scorers is larger than many regime-level differences, motivating multi-scorer reporting.}
\label{fig:app-nli-entailment-regime}
\end{figure}

Directional NLI is used as a sensitivity diagnostic for generated evidence. The premise is the reference evidence and the hypothesis is the generated evidence. This direction asks whether the generated evidence is supported by the reference evidence. Table~\ref{tab:app-nli} compares the four scorers. Figures~\ref{fig:app-nli-model-regime}--\ref{fig:app-nli-entailment-regime} then show scorer/regime variation, length sensitivity, and aggregate entailment rates. NLI does not prove medical correctness, and it can be sensitive to hypothesis length, generic evidence, and model calibration.

\begin{table*}[t]
\centering
\small
\begin{tabular}{lrrrrrrrr}
\toprule
NLI model & Runs & Rows & Ent. & Con. & Neu. & Overlap & DFS & Short \\
\midrule
BioLinkBERT-MedNLI & 150 & 1700.7 & 0.672 & 0.234 & 0.094 & 0.407 & 0.395 & 0.073 \\
PubMedBERT-MNLI-MedNLI & 150 & 1700.7 & 0.517 & 0.417 & 0.066 & 0.407 & 0.304 & 0.073 \\
DeBERTa-v3-ZS-NLI & 150 & 1700.7 & 0.245 & -- & -- & 0.407 & 0.149 & 0.073 \\
RoBERTa-large-MNLI & 150 & 1700.7 & 0.076 & 0.211 & 0.714 & 0.407 & 0.045 & 0.073 \\
\bottomrule
\end{tabular}
\caption{Directional NLI diagnostics over generated evidence. DeBERTa-v3 is binary entailment/not-entailment; contradiction and neutral are not reported.}
\label{tab:app-nli}
\end{table*}

NLI is directionally useful but should not be treated as definitive biomedical faithfulness. Models differ substantially in entailment propensity, and short/template-like evidence can receive inflated entailment scores.

\section{Generated Explanation Diagnostics}
\label{app:explanation-diag}

The generated explanation field is retained for transparency but is not treated as a headline faithfulness endpoint. As a lightweight output-consistency check, we measure explanation--verdict consistency: when an explanation explicitly restates a verdict, we compare that stated verdict with the system's parsed label. Table~\ref{tab:app-explanation-consistency} reports this diagnostic by regime. It detects internal self-contradiction between the structured output and the natural-language justification, but it does not prove that the explanation is medically faithful.

\begin{table}[t]
\centering
\small
\setlength{\tabcolsep}{2pt}
\begin{tabular}{lrrr}
\toprule
Regime & Expl.\ rate & Verdict-stated & Consistency \\
\midrule
Base LLM & 0.944 & 136{,}986 & 0.973 \\
RAG LLM & 0.983 & 39{,}770 & 0.970 \\
Fine-tuned LLM & 0.853 & 24{,}724 & 0.966 \\
\bottomrule
\end{tabular}
\caption{Explanation--verdict consistency by regime. \emph{Expl.\ rate} is the fraction of outputs with a non-empty explanation; \emph{Verdict-stated} counts explanations that explicitly restate a verdict; \emph{Consistency} is the fraction of those whose stated verdict matches the parsed label.}
\label{tab:app-explanation-consistency}
\end{table}

The diagnostic shows that explicit label--explanation self-contradiction is rare when explanations commit to a verdict. However, this is only an internal-consistency measure. It does not evaluate whether the explanation is supported by the generated evidence or by the CARE-XAI reference evidence. Explanation--evidence entailment is therefore left to the same cluster-based NLI pipeline used for generated-evidence diagnostics.

\section{Lexical Evidence Overlap and Scaling}

ROUGE and lexical overlap are retained as secondary diagnostics because biomedical evidence can be correct under paraphrase and incorrect under high lexical similarity. Table~\ref{tab:app-evidence-overlap} reports lexical overlap for generated evidence; classifier rows appear with zero overlap because they do not generate evidence. Table~\ref{tab:app-scaling} records the model-size rows used in the scaling analysis.

\begin{table*}[!t]
\centering
\small
\setlength{\tabcolsep}{2pt}
\begin{tabular*}{\textwidth}{@{\extracolsep{\fill}}lllrrrrr@{}}
\toprule
Reg. & Model & Prompt & Pairs & Tok. & Content & PredLen & GoldLen \\
\midrule
base & Gemma-4-12B & cot-fewshot & 1751 & 0.077 & 0.075 & 46.6 & 509.2 \\
base & Gemma-4-12B & pico-fewshot & 1752 & 0.076 & 0.073 & 46.5 & 509.3 \\
base & Gemma-4-12B & pico-zero-shot & 1751 & 0.079 & 0.076 & 53.2 & 509.5 \\
base & Gemma-4-12B & zero-shot & 1752 & 0.083 & 0.080 & 56.4 & 509.3 \\
base & Gemma-4-26B-A4B & cot-fewshot & 1748 & 0.080 & 0.079 & 51.7 & 509.7 \\
base & Gemma-4-26B-A4B & pico-fewshot & 1751 & 0.083 & 0.079 & 54.1 & 509.4 \\
base & Gemma-4-26B-A4B & pico-zero-shot & 1752 & 0.083 & 0.079 & 60.0 & 509.3 \\
base & Gemma-4-26B-A4B & zero-shot & 1752 & 0.089 & 0.084 & 64.2 & 509.3 \\
base & Gemma-4-31B & cot-fewshot & 1746 & 0.085 & 0.082 & 53.9 & 510.2 \\
base & Gemma-4-31B & pico-fewshot & 1752 & 0.087 & 0.082 & 55.1 & 509.3 \\
base & Gemma-4-31B & pico-zero-shot & 1752 & 0.079 & 0.077 & 54.0 & 509.3 \\
base & Gemma-4-31B & zero-shot & 1752 & 0.085 & 0.082 & 57.3 & 509.3 \\
base & Gemma-4-E2B & cot-fewshot & 1749 & 0.075 & 0.073 & 48.8 & 507.8 \\
base & Gemma-4-E2B & pico-fewshot & 1751 & 0.074 & 0.072 & 48.4 & 508.4 \\
base & Gemma-4-E2B & pico-zero-shot & 1751 & 0.082 & 0.075 & 55.4 & 508.4 \\
base & Gemma-4-E2B & zero-shot & 1751 & 0.089 & 0.081 & 59.5 & 508.4 \\
base & Gemma-4-E4B & pico-fewshot & 1752 & 0.076 & 0.072 & 47.7 & 509.3 \\
base & Qwen3.5-27B & pico-fewshot & 1752 & 0.107 & 0.096 & 77.9 & 509.3 \\
base & Qwen3.5-27B & pico-zero-shot & 1752 & 0.109 & 0.097 & 80.0 & 509.3 \\
base & Qwen3.5-27B & zero-shot & 1749 & 0.112 & 0.102 & 82.0 & 507.3 \\
base & Qwen3.5-2B & cot-fewshot & 1736 & 0.106 & 0.088 & 70.1 & 507.7 \\
base & Qwen3.5-2B & pico-fewshot & 1737 & 0.105 & 0.088 & 69.9 & 509.9 \\
base & Qwen3.5-2B & pico-zero-shot & 1746 & 0.120 & 0.097 & 83.1 & 507.8 \\
base & Qwen3.5-2B & zero-shot & 1750 & 0.117 & 0.095 & 80.3 & 509.7 \\
base & Qwen3.5-35B-A3B & cot-fewshot & 1737 & 0.116 & 0.102 & 83.1 & 510.8 \\
base & Qwen3.5-35B-A3B & pico-fewshot & 1736 & 0.110 & 0.096 & 79.8 & 509.9 \\
base & Qwen3.5-35B-A3B & pico-zero-shot & 1746 & 0.112 & 0.097 & 83.1 & 508.1 \\
base & Qwen3.5-35B-A3B & zero-shot & 1726 & 0.117 & 0.104 & 86.3 & 510.5 \\
base & Qwen3.5-4B & cot-fewshot & 1751 & 0.094 & 0.087 & 64.2 & 508.1 \\
base & Qwen3.5-4B & pico-fewshot & 1750 & 0.085 & 0.082 & 57.0 & 509.7 \\
base & Qwen3.5-4B & pico-zero-shot & 1752 & 0.103 & 0.091 & 77.0 & 509.3 \\
base & Qwen3.5-4B & zero-shot & 1745 & 0.106 & 0.095 & 76.3 & 508.5 \\
base & Qwen3.5-9B & cot-fewshot & 1749 & 0.105 & 0.093 & 73.5 & 509.1 \\
base & Qwen3.5-9B & pico-fewshot & 1751 & 0.102 & 0.091 & 71.5 & 509.1 \\
base & Qwen3.5-9B & pico-zero-shot & 1752 & 0.105 & 0.095 & 76.9 & 509.3 \\
base & Qwen3.5-9B & zero-shot & 1752 & 0.114 & 0.102 & 83.5 & 509.3 \\
base & Qwen3.6-27B & cot-fewshot & 1752 & 0.104 & 0.094 & 72.1 & 509.3 \\
base & Qwen3.6-27B & pico-fewshot & 1752 & 0.100 & 0.091 & 71.1 & 509.3 \\
base & Qwen3.6-27B & pico-zero-shot & 1751 & 0.099 & 0.090 & 71.7 & 509.0 \\
base & Qwen3.6-27B & zero-shot & 1752 & 0.105 & 0.097 & 76.6 & 509.3 \\
base & Qwen3.6-35B-A3B & cot-fewshot & 1725 & 0.110 & 0.099 & 77.7 & 508.2 \\
base & Qwen3.6-35B-A3B & pico-zero-shot & 1744 & 0.100 & 0.092 & 75.6 & 510.3 \\
base & Qwen3.6-35B-A3B & zero-shot & 1740 & 0.113 & 0.102 & 83.0 & 510.1 \\
FT & Gemma-4-31B & cot-fewshot & 1752 & 0.056 & 0.036 & 76.8 & 509.3 \\
FT & Gemma-4-31B & pico-fewshot & 1752 & 0.056 & 0.036 & 76.4 & 509.3 \\
FT & Gemma-4-E4B & cot-fewshot & 1735 & 0.073 & 0.069 & 42.9 & 506.8 \\
FT & Qwen3.5-35B-A3B & cot-fewshot & 1728 & 0.082 & 0.078 & 50.1 & 508.2 \\
FT & Qwen3.5-4B & cot-fewshot & 1742 & 0.068 & 0.069 & 40.8 & 509.8 \\
\bottomrule
\end{tabular*}
\caption{Evidence-overlap diagnostics for evidence-generating systems. These are lexical diagnostics, not semantic faithfulness scores (part 1 of 2).}
\label{tab:app-evidence-overlap}
\end{table*}

\begin{table*}[!t]
\centering
\small
\setlength{\tabcolsep}{2pt}
\begin{tabular*}{\textwidth}{@{\extracolsep{\fill}}lllrrrrr@{}}
\toprule
Reg. & Model & Prompt & Pairs & Tok. & Content & PredLen & GoldLen \\
\midrule
FT & Qwen3.5-4B & pico-fewshot & 1751 & 0.066 & 0.067 & 39.1 & 509.5 \\
FT & Qwen3.5-4B & zero-shot & 1719 & 0.087 & 0.080 & 54.2 & 509.1 \\
FT & Qwen3.6-27B & cot-fewshot & 1752 & 0.069 & 0.071 & 42.2 & 509.3 \\
FT & Qwen3.6-27B & pico-fewshot & 1752 & 0.072 & 0.073 & 43.8 & 509.3 \\
FT & Qwen3.6-27B & zero-shot & 1751 & 0.088 & 0.083 & 53.7 & 509.2 \\
FT & Qwen3.6-35B-A3B & cot-fewshot & 1750 & 0.078 & 0.074 & 48.2 & 509.1 \\
FT & Qwen3.6-35B-A3B & pico-zero-shot & 1752 & 0.087 & 0.077 & 55.9 & 509.3 \\
FT & Qwen3.6-35B-A3B & zero-shot & 1752 & 0.091 & 0.079 & 57.5 & 509.3 \\
rag & Gemma-4-26B-A4B & cot-fewshot & 1752 & 0.083 & 0.079 & 53.8 & 509.3 \\
rag & Gemma-4-26B-A4B & pico-fewshot & 1752 & 0.085 & 0.081 & 54.6 & 509.3 \\
rag & Gemma-4-26B-A4B & pico-zero-shot & 1752 & 0.101 & 0.093 & 70.6 & 509.3 \\
rag & Gemma-4-26B-A4B & zero-shot & 1751 & 0.102 & 0.095 & 69.7 & 509.5 \\
rag & Gemma-4-31B & cot-fewshot & 1751 & 0.088 & 0.083 & 57.9 & 509.5 \\
rag & Gemma-4-31B & pico-fewshot & 1752 & 0.086 & 0.082 & 57.0 & 509.3 \\
rag & Gemma-4-31B & pico-zero-shot & 1752 & 0.099 & 0.091 & 68.7 & 509.3 \\
rag & Gemma-4-31B & zero-shot & 1738 & 0.101 & 0.093 & 67.4 & 506.5 \\
rag & Gemma-4-E2B & cot-fewshot & 1751 & 0.085 & 0.076 & 55.9 & 508.4 \\
rag & Gemma-4-E2B & pico-fewshot & 1748 & 0.085 & 0.077 & 55.7 & 508.1 \\
rag & Gemma-4-E2B & pico-zero-shot & 1750 & 0.101 & 0.087 & 70.1 & 507.8 \\
rag & Gemma-4-E2B & zero-shot & 1750 & 0.107 & 0.091 & 72.7 & 508.3 \\
rag & Gemma-4-E4B & cot-fewshot & 1751 & 0.089 & 0.082 & 57.3 & 508.9 \\
rag & Gemma-4-E4B & pico-fewshot & 1752 & 0.087 & 0.080 & 57.6 & 509.3 \\
rag & Gemma-4-E4B & pico-zero-shot & 1752 & 0.102 & 0.091 & 72.5 & 509.3 \\
rag & Gemma-4-E4B & zero-shot & 1752 & 0.105 & 0.092 & 71.6 & 509.3 \\
rag & Qwen3.5-2B & cot-fewshot & 1730 & 0.098 & 0.085 & 65.8 & 510.6 \\
rag & Qwen3.5-2B & pico-fewshot & 1692 & 0.096 & 0.084 & 66.3 & 516.2 \\
rag & Qwen3.5-2B & pico-zero-shot & 1749 & 0.122 & 0.099 & 96.4 & 508.9 \\
rag & Qwen3.5-2B & zero-shot & 1752 & 0.120 & 0.100 & 89.3 & 509.3 \\
rag & Qwen3.5-35B-A3B & cot-fewshot & 1702 & 0.118 & 0.102 & 86.4 & 508.9 \\
rag & Qwen3.5-35B-A3B & pico-zero-shot & 1719 & 0.122 & 0.105 & 91.5 & 511.5 \\
rag & Qwen3.5-35B-A3B & zero-shot & 1740 & 0.124 & 0.107 & 92.2 & 511.2 \\
rag & Qwen3.5-4B & cot-fewshot & 1752 & 0.090 & 0.087 & 59.9 & 509.3 \\
rag & Qwen3.5-4B & pico-fewshot & 1752 & 0.095 & 0.089 & 64.6 & 509.3 \\
rag & Qwen3.5-4B & pico-zero-shot & 1730 & 0.106 & 0.096 & 76.3 & 504.9 \\
rag & Qwen3.5-4B & zero-shot & 1751 & 0.108 & 0.097 & 76.7 & 509.5 \\
rag & Qwen3.5-9B & cot-fewshot & 1752 & 0.102 & 0.094 & 71.2 & 509.3 \\
rag & Qwen3.5-9B & pico-fewshot & 1752 & 0.101 & 0.093 & 71.5 & 509.3 \\
rag & Qwen3.5-9B & pico-zero-shot & 1751 & 0.108 & 0.097 & 79.2 & 509.5 \\
rag & Qwen3.5-9B & zero-shot & 1752 & 0.109 & 0.100 & 79.5 & 509.3 \\
rag & Qwen3.6-27B & cot-fewshot & 1752 & 0.100 & 0.094 & 70.2 & 509.3 \\
rag & Qwen3.6-27B & pico-fewshot & 1752 & 0.102 & 0.095 & 71.9 & 509.3 \\
rag & Qwen3.6-27B & pico-zero-shot & 1752 & 0.107 & 0.099 & 76.5 & 509.3 \\
rag & Qwen3.6-27B & zero-shot & 1752 & 0.113 & 0.102 & 80.3 & 509.3 \\
rag & Qwen3.6-35B-A3B & cot-fewshot & 1729 & 0.111 & 0.099 & 80.9 & 508.6 \\
rag & Qwen3.6-35B-A3B & pico-fewshot & 1747 & 0.112 & 0.100 & 83.0 & 509.5 \\
rag & Qwen3.6-35B-A3B & pico-zero-shot & 1742 & 0.122 & 0.106 & 93.8 & 510.2 \\
rag & Qwen3.6-35B-A3B & zero-shot & 1710 & 0.129 & 0.109 & 98.4 & 512.7 \\
\bottomrule
\end{tabular*}
\caption{Evidence-overlap diagnostics for evidence-generating systems. These are lexical diagnostics, not semantic faithfulness scores (part 2 of 2).}
\label{tab:app-evidence-overlap-part-2}
\end{table*}

\begin{table*}[!t]
\centering
\small
\setlength{\tabcolsep}{3pt}
\begin{tabular*}{\textwidth}{@{\extracolsep{\fill}}lll llrr@{}}
\toprule
Family & Model & B & Reg. & Prompt & F1 & Acc. \\
\midrule
Gemma & Gemma-4-E2B & 2 & base & cot-fewshot & 0.365 & 0.368 \\
Gemma & Gemma-4-E2B & 2 & rag & cot-fewshot & 0.368 & 0.421 \\
Gemma & Gemma-4-E2B & 2 & base & pico-fewshot & 0.363 & 0.371 \\
Gemma & Gemma-4-E2B & 2 & rag & pico-fewshot & 0.376 & 0.434 \\
Gemma & Gemma-4-E2B & 2 & base & pico-zero-shot & 0.337 & 0.336 \\
Gemma & Gemma-4-E2B & 2 & rag & pico-zero-shot & 0.344 & 0.389 \\
Gemma & Gemma-4-E2B & 2 & base & zero-shot & 0.334 & 0.339 \\
Gemma & Gemma-4-E2B & 2 & rag & zero-shot & 0.295 & 0.376 \\
Gemma & Gemma-4-26B-A4B & 4 & base & cot-fewshot & 0.425 & 0.434 \\
Gemma & Gemma-4-26B-A4B & 4 & rag & cot-fewshot & 0.371 & 0.394 \\
Gemma & Gemma-4-E4B & 4 & rag & cot-fewshot & 0.368 & 0.400 \\
Gemma & Gemma-4-26B-A4B & 4 & base & pico-fewshot & 0.395 & 0.394 \\
Gemma & Gemma-4-E4B & 4 & base & pico-fewshot & 0.327 & 0.336 \\
Gemma & Gemma-4-26B-A4B & 4 & rag & pico-fewshot & 0.391 & 0.408 \\
Gemma & Gemma-4-E4B & 4 & rag & pico-fewshot & 0.388 & 0.416 \\
Gemma & Gemma-4-26B-A4B & 4 & base & pico-zero-shot & 0.391 & 0.396 \\
Gemma & Gemma-4-26B-A4B & 4 & rag & pico-zero-shot & 0.398 & 0.426 \\
Gemma & Gemma-4-E4B & 4 & rag & pico-zero-shot & 0.367 & 0.411 \\
Gemma & Gemma-4-26B-A4B & 4 & base & zero-shot & 0.402 & 0.416 \\
Gemma & Gemma-4-26B-A4B & 4 & rag & zero-shot & 0.383 & 0.411 \\
Gemma & Gemma-4-E4B & 4 & rag & zero-shot & 0.328 & 0.380 \\
Gemma & Gemma-4-12B & 12 & base & cot-fewshot & 0.375 & 0.385 \\
Gemma & Gemma-4-12B & 12 & base & pico-fewshot & 0.347 & 0.355 \\
Gemma & Gemma-4-12B & 12 & base & pico-zero-shot & 0.369 & 0.388 \\
Gemma & Gemma-4-12B & 12 & base & zero-shot & 0.411 & 0.431 \\
Gemma & Gemma-4-31B & 31 & base & cot-fewshot & 0.441 & 0.468 \\
Gemma & Gemma-4-31B & 31 & rag & cot-fewshot & 0.399 & 0.415 \\
Gemma & Gemma-4-31B & 31 & base & pico-fewshot & 0.415 & 0.414 \\
Gemma & Gemma-4-31B & 31 & rag & pico-fewshot & 0.409 & 0.423 \\
Gemma & Gemma-4-31B & 31 & base & pico-zero-shot & 0.414 & 0.422 \\
Gemma & Gemma-4-31B & 31 & rag & pico-zero-shot & 0.400 & 0.416 \\
Gemma & Gemma-4-31B & 31 & base & zero-shot & 0.417 & 0.429 \\
Gemma & Gemma-4-31B & 31 & rag & zero-shot & 0.396 & 0.411 \\
Qwen3.5 & Qwen3.5-2B & 2 & base & cot-fewshot & 0.335 & 0.382 \\
Qwen3.5 & Qwen3.5-2B & 2 & rag & cot-fewshot & 0.368 & 0.437 \\
Qwen3.5 & Qwen3.5-2B & 2 & base & pico-fewshot & 0.333 & 0.355 \\
\bottomrule
\end{tabular*}
\caption{Model-size rows used in the scaling analysis (part 1 of 3).}
\label{tab:app-scaling}
\end{table*}

\begin{table*}[!t]
\centering
\small
\setlength{\tabcolsep}{3pt}
\begin{tabular*}{\textwidth}{@{\extracolsep{\fill}}lll llrr@{}}
\toprule
Family & Model & B & Reg. & Prompt & F1 & Acc. \\
\midrule
Qwen3.5 & Qwen3.5-2B & 2 & rag & pico-fewshot & 0.361 & 0.403 \\
Qwen3.5 & Qwen3.5-2B & 2 & base & pico-zero-shot & 0.345 & 0.340 \\
Qwen3.5 & Qwen3.5-2B & 2 & rag & pico-zero-shot & 0.385 & 0.435 \\
Qwen3.5 & Qwen3.5-2B & 2 & base & zero-shot & 0.318 & 0.348 \\
Qwen3.5 & Qwen3.5-2B & 2 & rag & zero-shot & 0.320 & 0.423 \\
Qwen3.5 & Qwen3.5-35B-A3B & 3 & base & cot-fewshot & 0.416 & 0.442 \\
Qwen3.5 & Qwen3.5-35B-A3B & 3 & rag & cot-fewshot & 0.390 & 0.431 \\
Qwen3.5 & Qwen3.5-35B-A3B & 3 & base & pico-fewshot & 0.375 & 0.392 \\
Qwen3.5 & Qwen3.5-35B-A3B & 3 & base & pico-zero-shot & 0.410 & 0.431 \\
Qwen3.5 & Qwen3.5-35B-A3B & 3 & rag & pico-zero-shot & 0.400 & 0.438 \\
Qwen3.5 & Qwen3.5-35B-A3B & 3 & base & zero-shot & 0.444 & 0.471 \\
Qwen3.5 & Qwen3.5-35B-A3B & 3 & rag & zero-shot & 0.393 & 0.425 \\
Qwen3.5 & Qwen3.5-4B & 4 & base & cot-fewshot & 0.390 & 0.385 \\
Qwen3.5 & Qwen3.5-4B & 4 & rag & cot-fewshot & 0.365 & 0.414 \\
Qwen3.5 & Qwen3.5-4B & 4 & base & pico-fewshot & 0.366 & 0.364 \\
Qwen3.5 & Qwen3.5-4B & 4 & rag & pico-fewshot & 0.382 & 0.421 \\
Qwen3.5 & Qwen3.5-4B & 4 & base & pico-zero-shot & 0.351 & 0.354 \\
Qwen3.5 & Qwen3.5-4B & 4 & rag & pico-zero-shot & 0.380 & 0.421 \\
Qwen3.5 & Qwen3.5-4B & 4 & base & zero-shot & 0.418 & 0.429 \\
Qwen3.5 & Qwen3.5-4B & 4 & rag & zero-shot & 0.355 & 0.402 \\
Qwen3.5 & Qwen3.5-9B & 9 & base & cot-fewshot & 0.393 & 0.401 \\
Qwen3.5 & Qwen3.5-9B & 9 & rag & cot-fewshot & 0.401 & 0.434 \\
Qwen3.5 & Qwen3.5-9B & 9 & base & pico-fewshot & 0.357 & 0.374 \\
Qwen3.5 & Qwen3.5-9B & 9 & rag & pico-fewshot & 0.414 & 0.445 \\
Qwen3.5 & Qwen3.5-9B & 9 & base & pico-zero-shot & 0.415 & 0.427 \\
Qwen3.5 & Qwen3.5-9B & 9 & rag & pico-zero-shot & 0.413 & 0.460 \\
Qwen3.5 & Qwen3.5-9B & 9 & base & zero-shot & 0.431 & 0.459 \\
Qwen3.5 & Qwen3.5-9B & 9 & rag & zero-shot & 0.431 & 0.461 \\
Qwen3.5 & Qwen3.5-27B & 27 & base & cot-fewshot & 0.450 & 0.466 \\
Qwen3.5 & Qwen3.5-27B & 27 & rag & cot-fewshot & 0.407 & 0.434 \\
Qwen3.5 & Qwen3.5-27B & 27 & base & pico-fewshot & 0.417 & 0.418 \\
Qwen3.5 & Qwen3.5-27B & 27 & rag & pico-fewshot & 0.421 & 0.448 \\
Qwen3.5 & Qwen3.5-27B & 27 & base & pico-zero-shot & 0.426 & 0.434 \\
Qwen3.5 & Qwen3.5-27B & 27 & rag & pico-zero-shot & 0.399 & 0.422 \\
Qwen3.5 & Qwen3.5-27B & 27 & base & zero-shot & 0.450 & 0.483 \\
Qwen3.5 & Qwen3.5-27B & 27 & rag & zero-shot & 0.393 & 0.416 \\
\bottomrule
\end{tabular*}
\caption{Model-size rows used in the scaling analysis (part 2 of 3).}
\label{tab:app-scaling-part-2}
\end{table*}

\begin{table*}[!t]
\centering
\small
\setlength{\tabcolsep}{3pt}
\begin{tabular*}{\textwidth}{@{\extracolsep{\fill}}lll llrr@{}}
\toprule
Family & Model & B & Reg. & Prompt & F1 & Acc. \\
\midrule
Qwen3.6 & Qwen3.6-35B-A3B & 3 & base & cot-fewshot & 0.422 & 0.458 \\
Qwen3.6 & Qwen3.6-35B-A3B & 3 & rag & cot-fewshot & 0.391 & 0.431 \\
Qwen3.6 & Qwen3.6-35B-A3B & 3 & rag & pico-fewshot & 0.422 & 0.455 \\
Qwen3.6 & Qwen3.6-35B-A3B & 3 & base & pico-zero-shot & 0.419 & 0.444 \\
Qwen3.6 & Qwen3.6-35B-A3B & 3 & rag & pico-zero-shot & 0.405 & 0.453 \\
Qwen3.6 & Qwen3.6-35B-A3B & 3 & base & zero-shot & 0.437 & 0.481 \\
Qwen3.6 & Qwen3.6-35B-A3B & 3 & rag & zero-shot & 0.399 & 0.434 \\
Qwen3.6 & Qwen3.6-27B & 27 & base & cot-fewshot & 0.430 & 0.442 \\
Qwen3.6 & Qwen3.6-27B & 27 & rag & cot-fewshot & 0.390 & 0.420 \\
Qwen3.6 & Qwen3.6-27B & 27 & base & pico-fewshot & 0.408 & 0.409 \\
Qwen3.6 & Qwen3.6-27B & 27 & rag & pico-fewshot & 0.391 & 0.416 \\
Qwen3.6 & Qwen3.6-27B & 27 & base & pico-zero-shot & 0.419 & 0.426 \\
Qwen3.6 & Qwen3.6-27B & 27 & rag & pico-zero-shot & 0.411 & 0.440 \\
Qwen3.6 & Qwen3.6-27B & 27 & base & zero-shot & 0.434 & 0.470 \\
Qwen3.6 & Qwen3.6-27B & 27 & rag & zero-shot & 0.405 & 0.432 \\
\bottomrule
\end{tabular*}
\caption{Model-size rows used in the scaling analysis (part 3 of 3).}
\label{tab:app-scaling-part-3}
\end{table*}

The lexical-overlap tables should not be used to rank systems by faithfulness. They are useful for detecting degenerate evidence copying, empty evidence, and gross mismatch, but they cannot distinguish a faithful paraphrase from a hallucinated sentence with overlapping terminology. This is why the main paper emphasizes Bio-GRACE and why the human-verification protocol asks annotators to rate support, usefulness, and safety concern directly.

\section{Ensemble and Routing Headroom}

Table~\ref{tab:ensemble} compares majority voting, compact meta-votes, and a non-deployable cross-system oracle. It quantifies complementary errors and the maximum routing headroom available from the stored predictions.

\begin{center}
\small
\begin{tabular}{lllll}
\toprule
Method & Scope & Voters & F1 & Acc. \\
\midrule
Majority & Base & 44 & .409 & .414 \\
Meta-vote & Base & 3 & .422 & .428 \\
Majority & RAG & 43 & .396 & .426 \\
Meta-vote & RAG & 3 & .399 & .427 \\
Majority & FT & 13 & .475 & .536 \\
Meta-vote & FT & 3 & .477 & .542 \\
Majority & Clf. & 13 & .575 & .614 \\
Oracle & Cross & 3 & .618 & .658 \\
\bottomrule
\end{tabular}
\captionof{table}{Ensemble results; oracle is a non-deployable upper bound.}
\label{tab:ensemble}
\end{center}

The ensemble table is an oracle/headroom analysis rather than a deployable system. It estimates how much performance could improve if a selector knew which available system was correct for each item. This motivates routing and uncertainty work but is not used for headline claims.

The headroom analysis is useful because it shows that errors are not perfectly overlapping across systems. If a future selector could identify when fine-tuned LLMs, RAG systems, or classifiers are likely to be reliable, aggregate performance could improve without training a larger generator. The current paper does not claim such a selector has been learned; it only reports the available headroom.

\section{Uncertainty and Calibration}

\begin{figure}[t]
\centering
\begin{tikzpicture}
\begin{axis}[
  ybar,
  width=0.88\columnwidth,
  height=4.2cm,
  title={Classifier calibration error},
  ylabel={ECE},
  bar width=12pt,
  fill=teal!55,
  draw=teal!80!black,
  symbolic x coords={biobert-42,pubmedbert-2026,biobert-13,biobert-2026,pubmedbert-42,biolinkbert-2026},
  xtick=data,
  x tick label style={rotate=25,anchor=east,font=\scriptsize},
  ymin=0,
  enlarge x limits=0.16,
  grid=major,
  grid style={draw=gray!18},
  axis line style={draw=gray!50},
  tick style={draw=gray!50},
]
\addplot coordinates {(biobert-42,0.1129) (pubmedbert-2026,0.0361) (biobert-13,0.1286) (biobert-2026,0.0511) (pubmedbert-42,0.1113) (biolinkbert-2026,0.1136)};
\end{axis}
\end{tikzpicture}
\caption{Classifier calibration error for representative runs. Calibration is used as an uncertainty diagnostic rather than a final selection rule.}
\label{fig:app-calibration}
\end{figure}
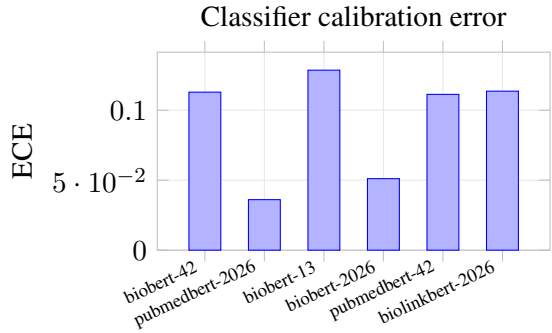

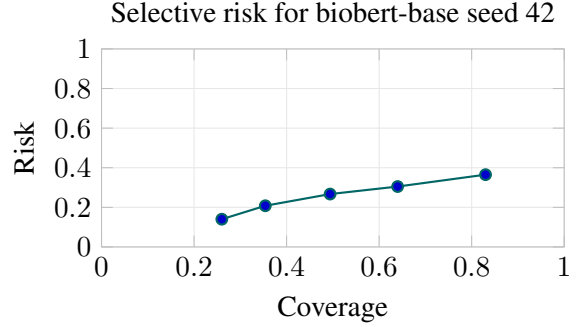
\begin{figure}[t]
\centering
\begin{tikzpicture}
\begin{axis}[
  width=\columnwidth,
  height=4.2cm,
  xlabel={Coverage},
  ylabel={Risk},
  title={Selective risk for biobert-base seed 42},
  xmin=0,xmax=1,ymin=0,ymax=1,
  grid=both,
  grid style={draw=gray!18},
  axis line style={draw=gray!50},
  tick style={draw=gray!50},
]
\addplot+[mark=*,teal!80!black,thick] coordinates {(0.830,0.365) (0.640,0.305) (0.494,0.267) (0.354,0.208) (0.260,0.140)};
\end{axis}
\end{tikzpicture}
\caption{Selective risk--coverage behavior for classifier confidence. Higher confidence improves risk on retained subsets but does not solve evidence generation.}
\label{fig:app-selective-risk}
\end{figure}

Table~\ref{tab:app-classifier-uncertainty} reports classifier entropy, confidence, margin, negative log-likelihood, Brier score, and expected calibration error. Figure~\ref{fig:app-calibration} compares calibration error, Figure~\ref{fig:app-selective-risk} shows risk--coverage behavior, and Figure~\ref{fig:app-llm-disagreement} gives the LLM disagreement proxy by source and label.

\begin{table*}[!t]
\centering
\small
\setlength{\tabcolsep}{1pt}
\begin{tabular*}{\textwidth}{@{\extracolsep{\fill}}llrrrrrrrrr@{}}
\toprule
Model & Seed & Rows & Acc. & F1 & Entropy & Conf. & Margin & NLL & Brier & ECE \\
\midrule
BioBERT-base & 13 & 1752 & 0.600 & 0.555 & 0.645 & 0.721 & 0.530 & 0.917 & 0.537 & 0.129 \\
BioBERT-base & 42 & 1752 & 0.604 & 0.560 & 0.647 & 0.714 & 0.519 & 0.924 & 0.536 & 0.113 \\
BioBERT-base & 2026 & 1752 & 0.573 & 0.550 & 0.801 & 0.621 & 0.385 & 0.850 & 0.512 & 0.051 \\
BioLinkBERT-large & 13 & 1752 & 0.588 & 0.541 & 0.649 & 0.710 & 0.505 & 0.894 & 0.524 & 0.122 \\
BioLinkBERT-large & 42 & 1752 & 0.587 & 0.538 & 0.633 & 0.719 & 0.518 & 0.900 & 0.532 & 0.132 \\
BioLinkBERT-large & 2026 & 1752 & 0.583 & 0.542 & 0.674 & 0.696 & 0.487 & 0.870 & 0.524 & 0.114 \\
PubMedBERT-base & 13 & 1752 & 0.575 & 0.536 & 0.807 & 0.621 & 0.383 & 0.831 & 0.499 & 0.056 \\
PubMedBERT-base & 42 & 1752 & 0.579 & 0.543 & 0.680 & 0.691 & 0.476 & 0.889 & 0.531 & 0.111 \\
PubMedBERT-base & 2026 & 1752 & 0.590 & 0.558 & 0.803 & 0.619 & 0.376 & 0.825 & 0.496 & 0.036 \\
\bottomrule
\end{tabular*}
\caption{Classifier uncertainty and calibration diagnostics.}
\label{tab:app-classifier-uncertainty}
\end{table*}

\begin{figure}[t]
\centering
\begin{tikzpicture}[x=1.05cm,y=0.45cm]
\node[anchor=east,font=\footnotesize] at (0,0) {};
\node[rotate=35,anchor=west,font=\footnotesize] at (1,0) {SUPPO.};
\node[rotate=35,anchor=west,font=\footnotesize] at (2,0) {CONTR.};
\node[rotate=35,anchor=west,font=\footnotesize] at (3,0) {UNADD.};
\node[anchor=east,font=\footnotesize] at (0,-1) {PubMedQA};
\fill[teal!38!white] (0.55,-1.35) rectangle (1.45,-0.65);
\node[font=\footnotesize] at (1,-1) {0.57};
\fill[teal!55!white] (1.55,-1.35) rectangle (2.45,-0.65);
\node[font=\footnotesize] at (2,-1) {0.90};
\fill[teal!46!white] (2.55,-1.35) rectangle (3.45,-0.65);
\node[font=\footnotesize] at (3,-1) {0.72};
\node[anchor=east,font=\footnotesize] at (0,-2) {SciFact};
\fill[teal!27!white] (0.55,-2.35) rectangle (1.45,-1.65);
\node[font=\footnotesize] at (1,-2) {0.36};
\fill[teal!45!white] (1.55,-2.35) rectangle (2.45,-1.65);
\node[font=\footnotesize] at (2,-2) {0.71};
\fill[teal!10!white] (2.55,-2.35) rectangle (3.45,-1.65);
\node[font=\footnotesize] at (3,-2) {0.00};
\node[anchor=east,font=\footnotesize] at (0,-3) {HealthVer};
\fill[teal!26!white] (0.55,-3.35) rectangle (1.45,-2.65);
\node[font=\footnotesize] at (1,-3) {0.33};
\fill[teal!40!white] (1.55,-3.35) rectangle (2.45,-2.65);
\node[font=\footnotesize] at (2,-3) {0.60};
\fill[teal!30!white] (2.55,-3.35) rectangle (3.45,-2.65);
\node[font=\footnotesize] at (3,-3) {0.40};
\node[anchor=east,font=\footnotesize] at (0,-4) {PUBHEALTH};
\fill[teal!38!white] (0.55,-4.35) rectangle (1.45,-3.65);
\node[font=\footnotesize] at (1,-4) {0.56};
\fill[teal!40!white] (1.55,-4.35) rectangle (2.45,-3.65);
\node[font=\footnotesize] at (2,-4) {0.60};
\fill[teal!40!white] (2.55,-4.35) rectangle (3.45,-3.65);
\node[font=\footnotesize] at (3,-4) {0.62};
\node[anchor=east,font=\footnotesize] at (0,-5) {HealthFC};
\fill[teal!30!white] (0.55,-5.35) rectangle (1.45,-4.65);
\node[font=\footnotesize] at (1,-5) {0.41};
\fill[teal!51!white] (1.55,-5.35) rectangle (2.45,-4.65);
\node[font=\footnotesize] at (2,-5) {0.82};
\fill[teal!46!white] (2.55,-5.35) rectangle (3.45,-4.65);
\node[font=\footnotesize] at (3,-5) {0.73};
\end{tikzpicture}
\caption{LLM vote-disagreement entropy by source and gold label. Disagreement is used as an epistemic proxy because token probabilities were not stored for LLM generations.}
\label{fig:app-llm-disagreement}
\end{figure}

Classifier confidence enables calibration and selective prediction analysis. LLM uncertainty is less direct because token probabilities were not retained, so we use model/prompt vote disagreement as an epistemic proxy. This proxy should be interpreted cautiously: it measures label disagreement, not semantic uncertainty over free-text evidence.

The selective-risk plot demonstrates that confidence can identify easier classifier cases, but it does not solve evidence generation. A high-confidence classifier verdict still lacks generated evidence. Conversely, high LLM agreement on a label does not guarantee that the generated evidence is complete or source-grounded. These diagnostics are best viewed as triage signals for human review or retrieval gating.

\section{Human Verification Protocol}

We prepared five practice items and 100 final items. The final sample contains 41 base, 26 RAG, and 33 fine-tuned evidence-generating outputs; 22 HealthVer, 22 PUBHEALTH, 22 HealthFC, 21 PubMedQA, and 13 SciFact examples; and 36 supported, 36 contradicted, and 28 unaddressed reference labels. It is intentionally enriched with disagreements, RAG regressions, incorrect verdicts, weak evidence, and suspected hallucinations, so its outcome rates are diagnostic rather than population estimates.

Two master's-level biotech annotators first read a one-page rubric and independently completed the five practice items. After discussing only those practice disagreements with the study supervisor, they annotated the 100 final items independently and did not discuss them with each other. Each sheet displayed the claim, trusted reference evidence, system verdict, and generated evidence, but hid the reference label, model, regime, prompt, and selection reason. Annotators did not search externally.

For each item, annotators marked (i) verdict correctness as \emph{yes}, \emph{no}, or \emph{uncertain}; (ii) evidence support as \emph{supported}, \emph{partially supported}, \emph{unsupported}, \emph{contradicted}, or \emph{unclear}; (iii) usefulness as \emph{useful}, \emph{partly useful}, or \emph{not useful}; and (iv) safety concern as \emph{none}, \emph{minor}, or \emph{major}, with optional notes. They were instructed to flag causal overstatement, changed certainty, population transfer, invented details, and omitted uncertainty. We report Cohen's $\kappa$ for the two humans, nominal Krippendorff's $\alpha$ across all four evaluators, exact-human-consensus outcome rates, and LLM agreement with that human consensus. LLM judges were independent sensitivity checks and did not adjudicate human disagreement.

\section{Failure Modes}

We retain failed and low-quality outputs in manifests instead of silently removing them. Important failure modes include invalid labels, empty evidence, template-copy outputs, retrieval-induced regressions, and plausible but unsupported biomedical details. Qwen3.5-9B fine-tuned outputs are excluded from performance claims due to zero valid-label coverage under strict parsing. Some reduced fine-tuned reruns remain incomplete because generation stopped before all 1,752 test rows were aligned.

The manifest policy is conservative. Planned but unavailable runs are reported as missing rather than deleted. Incomplete runs retain row-count and parse-coverage information but have null headline metrics. This prevents failed generations from being hidden while avoiding invalid metric comparisons.

Failure modes are especially important for evidence-generating systems because a malformed output can be more than an inconvenience. If a system emits an invalid label, omits evidence, or copies a template explanation, it cannot support biomedical fact-checking even if the underlying model sometimes performs well. The evaluation therefore treats output validity as part of system quality.

\section{Reproducibility Notes}

The evaluation excludes Mistral and Ministral models from paper-facing summaries, preserves raw model-output directories, and treats incomplete experiments as rows with null metrics. Non-RAG baseline duplicate directories were collapsed because corresponding predictions were byte-identical. Retrieval traces were frozen before downstream evaluation, and leakage-safe summaries were computed as sensitivity analyses. Training and inference used the NI-HPC cluster with AMD Instinct MI300X accelerators (192~GB HBM per GPU). Aggregate compute is estimated at approximately 10--15 GPU-days (240--360 GPU-hours), including completed and failed runs; exact accounting was not retained. The anonymized artifact records available scripts, prompts, structured outputs, and evaluator inputs.

\section{Human and LLM Verification Details}
\label{app:human-verification}

The verification sample contains 100 blinded outputs stratified across base, RAG, and fine-tuned regimes and enriched with difficult cases. Two biotech students (H1 and H2) and two independent LLM judges (L1 and L2) applied the same rubric. The LLM judgments were collected independently and were not used to adjudicate human disagreements. Because the sample deliberately emphasizes difficult outputs, the following rates are diagnostic rather than population-level estimates.

\paragraph{Pairwise agreement.}
Table~\ref{tab:pairwise-human-llm} reports pairwise agreement among all four evaluators. Agreement varies considerably by evaluator and rubric field. In particular, L2 shows very low verdict agreement with both human annotators but higher agreement for usefulness and safety.

\begin{table*}[t]
\centering
\small
\begin{tabular}{lrrrr}
\toprule
Pair & Verdict & Support & Usefulness & Safety \\
\midrule
H1--H2 & .265 (.69) & .391 (.56) & .155 (.48) & .250 (.62) \\
H1--L1 & .222 (.62) & .303 (.48) & .271 (.54) & .169 (.53) \\
H1--L2 & .049 (.15) & .270 (.37) & .379 (.61) & .327 (.72) \\
H2--L1 & .560 (.80) & .293 (.47) & .415 (.77) & .334 (.61) \\
H2--L2 & .104 (.21) & .244 (.35) & .249 (.62) & .231 (.64) \\
L1--L2 & .145 (.31) & .230 (.36) & .353 (.64) & .210 (.56) \\
\bottomrule
\end{tabular}
\caption{Pairwise evaluator agreement. Each cell reports Cohen's $\kappa$, with raw agreement in parentheses. H1 and H2 are the biotech annotators; L1 and L2 are the independent LLM judges.}
\label{tab:pairwise-human-llm}
\end{table*}

\paragraph{Multi-rater agreement.}
As shown in Table~\ref{tab:four-rater-agreement}, four-rater reliability is low across all dimensions. Strict three-of-four majorities are available for only 45\% of evidence-support ratings, although coverage is higher for usefulness and safety.

\begin{table*}[t]
\centering
\small
\begin{tabular}{lrrr}
\toprule
Field & $\alpha$ & Unanimous & Majority \\
\midrule
Verdict correctness & .105 & 14\% & 63\% \\
Evidence support & .263 & 22\% & 45\% \\
Evidence usefulness & .276 & 37\% & 68\% \\
Safety concern & .239 & 35\% & 73\% \\
\bottomrule
\end{tabular}
\caption{Krippendorff's nominal $\alpha$, unanimous agreement, and strict three-of-four majority coverage across the four evaluators.}
\label{tab:four-rater-agreement}
\end{table*}

\paragraph{Agreement with human consensus.}
For a more defensible comparison, Table~\ref{tab:llm-human-consensus} evaluates each LLM judge only on items for which H1 and H2 gave exactly the same rating. L1 aligns more strongly with human verdict judgments, while both LLMs align comparatively well on usefulness. L2 agrees more strongly with humans on safety than on verdict correctness.

\begin{table*}[t]
\centering
\small
\begin{tabular}{lrrrrr}
\toprule
 & & \multicolumn{2}{c}{L1} & \multicolumn{2}{c}{L2} \\
\cmidrule(lr){3-4}\cmidrule(lr){5-6}
Field & $n$ & Agr. & $\kappa$ & Agr. & $\kappa$ \\
\midrule
Verdict correctness & 69 & .855 & .653 & .203 & .107 \\
Evidence support & 56 & .571 & .389 & .429 & .332 \\
Evidence usefulness & 48 & .854 & .645 & .854 & .657 \\
Safety concern & 62 & .645 & .304 & .839 & .474 \\
\bottomrule
\end{tabular}
\caption{Agreement of each LLM judge with exact human consensus. The number of eligible items varies because human agreement differs across fields.}
\label{tab:llm-human-consensus}
\end{table*}

\paragraph{Consensus results by regime.}
Table~\ref{tab:human-regime-results} reports rubric outcomes using only exact H1--H2 agreement. Fine-tuned outputs have the highest agreed verdict-correctness rate, whereas RAG outputs have higher evidence-support and usefulness rates among the items with consensus. These differences should be interpreted cautiously because the denominators are small, vary by field, and arise from an intentionally enriched sample.

\begin{table*}[t]
\centering
\small
\begin{tabular}{lrrrrr}
\toprule
Regime & $N$ & Verdict & Support & Useful & Safety \\
\midrule
Base & 41 & 13.8 (29) & 9.5 (21) & 7.1 (14) & 24.1 (29) \\
RAG & 26 & 15.8 (19) & 42.9 (14) & 37.5 (16) & 8.3 (12) \\
Fine-tuned & 33 & 28.6 (21) & 14.3 (21) & 16.7 (18) & 23.8 (21) \\
\bottomrule
\end{tabular}
\caption{Human-consensus results by regime. Values are percentages; the number of fields with exact human agreement is shown in parentheses. Support combines \emph{supported} and \emph{partially supported}; Useful combines \emph{useful} and \emph{partly useful}; Safety denotes any minor or major concern.}
\label{tab:human-regime-results}
\end{table*}

These results reinforce the distinction between verdict accuracy and evidence quality. A system may produce the correct label while generating evidence that is unsupported or unusable. The low multi-rater agreement also demonstrates that automated LLM judging remains sensitive to judge choice. Accordingly, we report the human-consensus results as primary, use LLM agreement as a sensitivity analysis, and leave human disagreements unresolved rather than allowing an LLM to determine the final label.

\end{document}